\documentclass{article}

\usepackage[letterpaper, portrait, margin=1in]{geometry}

\usepackage[utf8]{inputenc}
 
\usepackage{balance}

\usepackage{amsmath,amssymb,amsfonts}
\usepackage{float}
\usepackage{subfig}
\usepackage{multirow}
\usepackage[T1]{fontenc}
\PassOptionsToPackage{hyphens}{url}\usepackage{hyperref}
\usepackage{url}

\usepackage{graphicx}
\usepackage{booktabs}
\usepackage{hhline}

\newcommand{\descr}[1]{\vspace{0.2cm} \noindent \textbf{#1}}

\usepackage{xcolor, colortbl}
\definecolor{citegreen}{HTML}{458B00}
\usepackage{hyperref}
\hypersetup{
   colorlinks=true,
   citecolor=citegreen
}

\newif\ifcomment
\commenttrue
\ifcomment

    \newcounter{GTNumberOfComments}
    \stepcounter{GTNumberOfComments}
    \newcommand{\gtnote}[1]{\textcolor{blue}{\small \bf [GT: \#\arabic{GTNumberOfComments}\stepcounter{GTNumberOfComments}: #1]}}
    
    \newcounter{SKNumberOfComments}
    \stepcounter{SKNumberOfComments}
    \newcommand{\sknote}[1]{\textcolor{magenta}{\small \bf [SK: \#\arabic{SKNumberOfComments}\stepcounter{SKNumberOfComments}: #1]}}

    \newcommand{\NOTE}[1]
    {
      {\footnotesize\it
        \begin{center}
          \begin{tabular}{|c|}
           \hline
            \parbox{0.85\columnwidth}{
              \medskip
              #1
              \medskip} \\
            \hline
          \end{tabular}
        \end{center}
        }
    }
\else
      \newcommand\gtnote[1]{}
      \newcommand\sknote[1]{}
    \newcommand\NOTE[1]{}
\fi

\begin{document}

\title{On the Adversarial Inversion of Deep Biometric Representations}



\author{Gioacchino Tangari$^1$, Shreesh Keskar$^{1, 2}$, Hassan Jameel Asghar$^1$ \& Dali Kaafar$^1$ \\
\small $^1$Macquarie University, Australia\\
\small $^2$BITS Pilani, India\\
\small \texttt{ninnitangari@gmail.com}\\
\small \texttt{f20150477g@alumni.bits-pilani.ac.in}\\ 
\small \texttt{\{hassan.asghar, dali.kaafar\}@mq.edu.au}
}

\maketitle





\begin{abstract}{
Biometric authentication service providers often claim that it is not possible to reverse-engineer a user's raw biometric sample, such as a fingerprint or a face image, from its mathematical (feature-space) representation. This is presented as a security feature of the system against an attacker who may be able to retrieve the template in the feature space. In this paper, we investigate this claim on the specific example of deep neural network (DNN) embeddings. Inversion of DNN embeddings has been investigated for explaining deep image representations or synthesizing normalized images. When setting the inversion, existing studies leverage full access to all layers of the original model, as well as all possible information on the original dataset -- including, in some cases, the original image to be reconstructed. For the biometric authentication use case, we need to investigate this under adversarial settings where an attacker has access to a feature-space representation but no direct access to the exact original dataset nor the original learned model. Instead, we assume varying degree of attacker's background knowledge about the distribution of the dataset as well as the original learned model (architecture and training process). In the worst case, we assume attacker has no knowledge of the original data distribution nor the original model. In these cases, we show that the attacker can exploit off-the-shelf DNN models and public datasets, to mimic the behaviour of the original learned model to varying degrees of success, based only on the obtained representation and attacker's prior knowledge. We propose a two-pronged attack that first infers the original DNN by exploiting the model footprint on the embedding, and then reconstructs the raw data by using the inferred model. We show the practicality of the attack on popular DNNs trained for two prominent biometric modalities, face and fingerprint recognition. The attack can effectively infer the original recognition model (mean accuracy 83\% for faces, 86\% for fingerprints), and can craft effective biometric reconstructions that are successfully authenticated with 1-vs-1 authentication accuracy of up to 92\% for some models.
}
\end{abstract}

\pagebreak

\tableofcontents

\pagebreak

\section{Introduction}

Deep Neural Networks (DNNs) have recently emerged as 
a great fit for critical applications on biometric data (e.g., Apple Face ID~\cite{applenn}~\cite{face-id-doc}), due to low error rates and competitive verification times~\cite{survey2018}~\cite{survey2019}. 
This is due to the ability of DNNs to map complex input (e.g., face image) into numerical representations (\textit{embeddings}) that, despite their compact size (e.g., 128 bytes per face~\cite{facenet}, 200 bytes per fingerprint~\cite{engelsma19}), can retain the most discriminative features about a subject. 
Our interest in deep embeddings is partly motivated by a biometric authentication use case. Current biometric technologies such as Microsoft's Windows Hello~\cite{win-hello} and Apple's Touch~\cite{touch-id} and Face ID~\cite{face-id} store biometric templates in the feature space (as embeddings) rather than in its original form (facial or fingerprint image). This is intriguingly flagged as a security feature, by arguing that an attacker cannot retrieve the original biometric sample from this \emph{mathematical representation}. The following quotes serve as examples:

\begin{itemize}
    \item ``Touch ID doesn't store any images of your fingerprint, and instead relies only on a mathematical representation. It isn't possible for someone to reverse engineer your actual fingerprint image from this stored data.''~\cite{touch-id}.
    \item ``Additionally, even if an attacker was actually able to get the biometric data from a device, it cannot be converted back into a raw biometric sample that could be recognized by the biometric sensor.''~\cite{win-hello}  
    \item ``Face images captured during normal operation aren’t saved, but are instead immediately discarded once the mathematical representation is calculated for either enrollment or comparison to the enrolled Face ID data.''~\cite{face-id-doc}.
    
\end{itemize}




A natural question is: given the feature-representation of a biometric trait,
can the attacker retrieve the original raw sample, e.g., a face image or a fingerprint, or at least gain some information about it, \emph{without} access to the biometric service? This is not merely of theoretical interest and relates to real-world scenarios. For instance, on-device biometric templates may be stolen or intercepted on a mobile phone infected by malware~\cite{kaspersky}.\footnote{Biometric template are not kept perpetually encrypted since template matching needs to account for small errors caused by intra-subject variations. Current technology does this in the unencrypted domain.} Crucially, attackers may maliciously obtain biometric features without having access to the models that originally generated them, as stored biometric templates of target users can be exposed in hacked databases~\cite{equifax}. Likewise, in online biometric authentication systems, a database breach exposes biometric templates of many users, as was reported in the case of Biostar 2 where 23GB of biometric data was leaked~\cite{biostar2}.

Since DNNs are increasingly being used for biometric recognition~\cite{survey2018}~\cite{survey2019}, including in proprietary systems, e.g., Apple's Face ID~\cite{applenn}, we restrict our focus to deep embeddings. There are many reasons for the popularity of DNNs in biometric recognition, including the widely available and highly accurate pre-trained models such as Facenet~\cite{facenet}, which means service providers do not need to train a DNN from scratch which requires the expensive process of diverse large-scale data collection (cf. Section~\ref{sec::advscenario}). The inversion of DNN embeddings has been an active research topic in computer vision, with the objective to improve visualization of DNNs~\cite{yosinski2015understanding}, to better explain their embeddings~\cite{MahendranV15}, or to normalize their input data~\cite{Cole2017SynthesizingNF}. These studies assume access to all layers of the original DNN model (e.g., autoencoder setup in ~\cite{dosovitskiy2015inverting}), and they leverage full knowledge of the model's training data (including, in \cite{mahendran2016visualizing} and \cite{yosinski2015understanding}, the original image).

In this paper, we investigate the inversion of deep biometric representations in an \textit{adversarial} scenario where the attacker does not have direct access to the target DNN model but has varying degrees of knowledge of (i) 
the biometric data distribution; and (ii) the target DNN model. 
We show that the attacker can at least partially infer the target model by exploiting the DNN \textit{footprint} on the embeddings, and can then apply this knowledge to solve the (otherwise ill-posed) inversion problem 
and craft biometric reconstructions.

In summary, we make the following contributions:

\begin{itemize}
    \item ~We propose a two-phase embedding inversion attack: the first phase, {\it Model Inference}, uses the observed embedding to predict the target (DNN) model; the second phase, {\it Reconstruction}, leverages an approximation of the real target model, together with an adversarial dataset, to train a biometric reconstructor. 
    \item ~For model inference, we experimentally demonstrate that, in DNN-based biometric recognition, it is possible to invert the embedding even if target model is not initially given to the adversary. This capability relies on the information {\it leaked} by feature vectors about the original DNN. We show that the attacker can effectively predict the original model, even when the attacker does not have \textit{similar} data (e.g., similarly pre-processed) as the biometric system's data (accuracy range [78\% - 99\%]), and also when the original DNN is fine-tuned towards a specific recognition task (accuracy range [60\% - 72\%]).
    \item ~For reconstruction, we formulate two practical adversarial scenarios where the attacker can approximate the target model and apply this knowledge to recover the original biometric trait. The scenarios build on the recent trend to \textit{re-purpose} pre-trained, off-the-shelf DNNs towards a specific biometric recognition task (generally, through additional training), which is the case for many biometric modalities, including fingerprint~\cite{fingernet}~\cite{michelsanti2017fast}, ear~\cite{ear}, palmprint~\cite{palmprint}, face~\cite{facetl}. 
    \item ~We test the attack on {\em face} and {\em fingerprint} recognition, by experimenting with 10 different DNN models and 5 biometric datasets. The attack produces reconstructions that are {\em close} to the originals (\textit{i.e.,} retaining most of face semantics, fingerprint details) when (i) the attacker has none or very limited knowledge on the biometric data, and (ii) the attacker can only {\em approximate} the real target model. The obtained reconstructions are also effective for impersonation, with up to 92\% samples accepted in (1-vs-1) verification, and up to 73\% correct identity classifications (1-out-of-$n$) -- 0.3\% being the success rate of a random guess. Notably, average success rates are still significant when the attacker has no prior knowledge on the biometric data ($\ge 20\%$), or can only recover a partial view on the target model ($\ge 25\%$). 
    \item For our attack, we assume a spectrum of background knowledge available to the adversary. This includes the case when the target model is built on top of some pre-trained model from a pool of prominent public DNN architectures available to the attacker, and varying knowledge about the target dataset, e.g., other data samples (fingerprints or face images) from the same subject (one whose template is acquired by the attacker), or data samples from completely different subjects (available as public data). Even in the worst case, i.e., the target model not being in the attackers pool of possible DNN architectures, and no samples from the target subject, we show that the attack performs significantly better than random guess to impersonate the subject (Section~\ref{sec:lim}).
\end{itemize}

While we specifically address the biometric use case, our inversion attack can be extended to other image tasks using DNNs (beyond biometrics), and in principle also to the text domain. Deep embeddings from pre-trained DNNs are, in fact, also becoming the standard in many natural language processing tasks, and have been shown to leak information on the input data~\cite{song2020information}. 

\section{Related work}

\descr{Inversion of deep representations.} Despite the impressive results achieved by DNNs in many domains, including biometric recognition~\cite{survey2018}~\cite{survey2019}, it is still unclear why they perform so well. One method to improve the understanding and the ``explainability" of DNNs is by inverting DNN embeddings~\cite{MahendranV15}~\cite{nash2019inverting}~\cite{du2018towards}~\cite{saad2007neural}~\cite{yosinski2015understanding}~\cite{jacobsen2018revnet}~\cite{mahendran2016visualizing}. The initial approach by Mehendran and Vedaldi~\cite{MahendranV15} works by setting one optimization problem for each embedding to invert. However, this technique yields limited accuracy due to the use of hand-crafted image priors~\cite{yosinski2015understanding} (e.g., Total Variation~\cite{MahendranV15}). Differently, the approach in ~\cite{jacobsen2018revnet} and \cite{nash2019inverting} consists in training an inversion model (generally, a DNN), thus it is closer in spirit to ours. A similar ``training-based" approach is also used by Cole et al. in ~\cite{Cole2017SynthesizingNF} to synthesize normalized images. All these studies, including training-based inversion ~\cite{jacobsen2018revnet}~\cite{nash2019inverting}~\cite{yosinski2015understanding}, require access to all layers of the original model, and leverage full knowledge of the model training data. The work closest to ours is the one from Mai et al.~\cite{mai2018reconstruction}. Two main differences between their work and our is (a) they assume that the attacker has (black-box) access to the target feature extractor, and (b) their analysis is limited to face templates, and does not include fingerprints. In contrast, in our adversarial settings, the attacker (i) cannot access the real data of the biometric system and (ii) does not know the original DNN beforehand. 

\descr{Model inversion attacks.} Recent research has shown that ML models (including DNNs) leak non-trivial information about their training data. Model Inversion attacks~\cite{fried2014}~\cite{fredrikson15}~\cite{ccs19}~\cite{zhang2020secret}, in particular,  try to \textit{invert} a ML model to (i) infer sensitive attributes of training data or to (ii) reproduce a plausible sample of a given training class. Both the work by Fredrikson et al.~\cite{fredrikson15} and the more recent one by Yang et al.~\cite{ccs19} implement the latter case as a \textit{face reconstruction} attack, where the adversary reconstructs the victim's face from the ML model output. In \cite{fredrikson15}, the attack is set against  both shallow and deep ML models in the form of optimization, regularized with denoising and sharpening manipulation. To overcome the problem of hand-crafted image priors, the attacks in \cite{ccs19}~\cite{zhang2020secret} train an adversarial model to invert a neural-network classifier. There are two key differences between Model Inversion and our scenario. \textit{First}, whereas Model Inversion’s problem is to find a realistic sample  labelled  in  a  specific  way  by  a  classifier,  our attack looks for the most probable data represented in a given way in the \textit{latent space} of a deep neural network. \textit{Second}, Model Inversion assumes black-box access (through query API~\cite{ccs19}~\cite{fredrikson15}) or white-box access (auto-encoder training in \cite{ccs19}, aggregate knowledge on training set in \cite{fredrikson15}). In our scenario, in contrast, the attacker is not given any knowledge of the target model or access to it.

\descr{Model stealing \& reverse-engineering.} Several studies~\cite{tramer2016stealing}~\cite{iclr2018}~\cite{orekondy2019knockoff}~\cite{wang2018stealing}~\cite{duddu2018stealing} have investigated how to recover attributes of an unknown ML model to shed light on its internals (reverse-engineering) or steal them. Existing \textit{model stealing} attacks attempt to infer model parameters~\cite{tramer2016stealing}~\cite{duddu2018stealing}, hyperparameters~\cite{wang2018stealing}~\cite{iclr2018}, decision boundaries~\cite{athalye2018obfuscated}, functionality~\cite{orekondy2019knockoff}. 
These approaches require black-box access to the model, and thus they cannot fit our settings, in which the target model (biometric feature extractor) cannot be queried as it only ``manifests" itself via one feature vector. 

\section{Threat Model and Adversarial Scenarios} 
\label{sec::advscenario}

We consider an adversary who maliciously obtains a biometric feature vector (\textit{embedding}). The embedding is generated by the original DNN model (\textit{target model}), extracting biometric features as part of the recognition system. 

\descr{Adversary's goal.} The adversary's objective is to recover, from the observed embedding, the original biometric information of a subject, e.g., someone's face or fingerprint.
Formally, the attacker wants to find the estimate $x^*$ of the real biometric trait satisfying the following:
\begin{equation}
\begin{aligned}
\label{eq:1}
x^* = \underset{x}{\text{argmax }} p_X(x) \text{ s.t. } \Phi(x)=y
\end{aligned}
\end{equation}
where $p_X$ is the probability distribution of the data (e.g., distribution of face images), $y$ is the embedding that the attacker observes and wants to {\em invert}, and $\Phi$ is the feature extractor that generated $y$, \textit{i.e.}, the target DNN model.  The ultimate goal can be to reveal or trade sensitive biometric information belonging to specific users or companies (e.g., Biostar2~\cite{biostar2}), or to impersonate the
subject in the target biometric recognition system.

\textit{Remark:} Equation~\ref{eq:1} resembles the formal definition of model inversion attacks~\cite{fredrikson15}~\cite{ccs19}, with the key difference that $\Phi$ is a feature extractor instead of a supervised classifier. Hence, whereas model inversion's goal is to find the most probable data labelled in a specific way by a classifier, in our settings the adversary looks for the most probable data represented as $y$ in the \textit{latent space} of a DNN model. 




\descr{Attack setting, assumptions and differences from prior work.}
Compared to embedding inversion in a ``controlled" environment \cite{dosovitskiy2015inverting}~\cite{MahendranV15}, which leverages full knowledge of original model and data, our adversarial inversion introduces two key challenges:

\textbullet ~{\em First,} the adversary has no access to the original (biometric) data; specifically, we assume that the attacker only knows the biometric modality (e.g., face, fingerprint) corresponding to the embedding in advance. {However, the distribution of the adversary's data, as represented by the identities and the fidelity/pre-processing of biometric images, may overlap with the original data distribution to a varying extent. This may be due to prior knowledge possessed by the adversary or due to the adversarial dataset being extensive enough to subsume some of the original data distribution.}

\textbullet ~{\em Second,} the attacker also does not know a-priori the model $\Phi$ that generated the embedding, nor is given ``query" access (e.g., feature-extraction API) to it. However, we assume that the biometric recognition builds on a pre-trained model of a \textit{prominent} DNN architecture (e.g., ResNets~\cite{resnet}, VGG~\cite{vgg}), available in deep-learning repositories like Tensorflow Hub~\cite{tfhub}, ModelZoo~\cite{modelzoo}, Open Neural Network Exchange~\cite{onnx} or public code (e.g., \textit{github}) repositories. The initial pre-trained model is then extended or adjusted to specific biometric task and data.

\emph{Justification:} This assumption is backed by the recent trend of fine-tuning pre-trained, publicly available DNNs towards biometric recognition tasks~\cite{survey2018}~\cite{survey2019}, which is gaining popularity for all main biometric modalities, including fingerprint~\cite{fingernet}~\cite{michelsanti2017fast}, iris~\cite{iris}, face~\cite{facetl}, palmprint~\cite{palmprint}, ear~\cite{ear}.\footnote{One exception is voice, due to the different nature of the data~\cite{survey2019}.} The reason behind the trend is two-fold: (i) the availability of pre-trained, off-the-shelf DNN models, including not only generic recognition models (e.g., ImageNet), but also biometric-specific models with state-of-the-art performance (e.g., Facenet~\cite{facenetcode}); (ii) for many biometric modalities, the unavailability of large-scale datasets (e.g., for ear recognition) or datasets with many samples per identity (as in the case of fingerprints~\cite{fingernet}) prevents training a DNN from scratch or making it converge to good local minimum~\cite{survey2019}. 

Specifically, we formulate two scenarios based on how the initial DNN model is used for biometric recognition:



\textbullet ~\textit{Scenario 1: Vanilla feature extractor} -- In this scenario, the initial pre-trained DNN 
is employed \textit{as is} to extract the embedding from the biometric input. 
Therefore, if the attacker can \textit{guess} the initial pre-trained DNN, the attacker would become able to generate feature vectors equal to the biometric system's ones. In other words, if attacker's guess is correct, it would perfectly match $\Phi$ from Equation~\ref{eq:1}.


\textbullet ~\textit{Scenario 2: Adapted feature extractor} -- In this scenario, the whole pre-trained DNN, or part of it, is fine-tuned to specialize on the biometric recognition task. 
In this case, by correctly guessing the initial DNN, the attacker can only obtain an approximation of the real feature extractor $\Phi$.

In both scenarios, the attacker's data ``resembles'' the original data to a certain extent, as detailed in Section~\ref{sec:attackcond}. 

\section{Approach}
The attack starts from the observation of a single embedding, which is the only ``manifestation" of the original DNN model -- we also call it \textit{target model}. To address such scenario, we propose a novel, two-phase approach.
In the first phase, called \textit{Model Inference}, the attacker infers the target model by exploiting the ``footprint" left by the original DNN model on feature vectors. In the second phase, called \textit{Reconstruction}, the attacker sets a reconstruction task to transform the embedding back into a biometric trait. 

\subsection{Model Inference Phase} 

Model Inference's goal is to infer from the observation of one biometric feature vector (deep embedding), the \textit{target model} that originally generated it. Our initial observation is that embeddings leak information on the target model. To assess the type and amount of model information an embedding reveals, we tested the inference of architecture (and training) model characteristics on 5,000 image recognition models from MNIST-NET~\cite{mnistnet} -- a large-scale dataset of diverse MNIST~\cite{lecun2010mnist} models. Specifically, we train a classifier which, for each embedding, outputs several model attributes. We show the results for a subset of attributes from~\cite{iclr2018} in Table~\ref{tab:motex} and compare it against Kennen~\cite{iclr2018}, which in contrast infers model characteristics using the target model's prediction outputs. Accuracy is $>$90\% for all tested embedding lengths, 
indicating strong inference power of embeddings on the target model.

\begin{table*}[!t]
\caption{Prediction of model characteristics from individual embeddings (Emb.) of different length (300, 500, 1000). Baseline is the ``Kennen" approach~\cite{iclr2018}, which instead uses 1000 probabilities from the model's prediction output.}
\resizebox{\linewidth}{!}{%
\begin{tabular}{l lllllllll}
&  \multicolumn{8}{l}{\textbf{Prediction Accuracy (\%)}} \\ \midrule
& \textit{Activation} & \textit{Dropout} & \textit{Max Pooling} & \textit{N. Conv. Layers} & \textit{Conv. Kernel Size} & \textit{N. Dense Layers} & \textit{Opt. Alg.} & \textit{Batch Size} \\ 
 & [ReLU,PReLU & [Yes,No] & [Yes,No] & [2,3,4] & [3x3, 5x5] & [2,3,4] & [SGD,Adam & [64,128,256] \\
  &  ELU,Tanh] &  &  &  &  & & RMSprop] &  \\\midrule
Emb.300  &  91.6 & 97.9 & 99.3 & 94.0 & 99.6 & 91.5 & 92.7 & 91.6  \\
Emb.500 & 93.8 & 97.9 & 99.5 & 96.7 & 99.7 & 93.8 & 94.3 & 94.5  \\
Emb.1000  & 96.3 & 98.3 & 99.5 & 96.5 & 99.6 & 95.4 & 96.4 & 95.4  \\
Kennen~\cite{iclr2018} & 80.6 & 94.6 & 94.9 & 67.1 & 84.6 & 77.3 & 71.8 & 50.4  \\ \midrule
Chance & 25.0 & 50.0 & 50.0 & 33.3 & 50.0 & 33.3 & 33.3 & 33.3 \\
\bottomrule
\end{tabular}
}
\label{tab:motex}
\end{table*}

 \begin{figure}[t]
  \includegraphics[width=1\columnwidth]{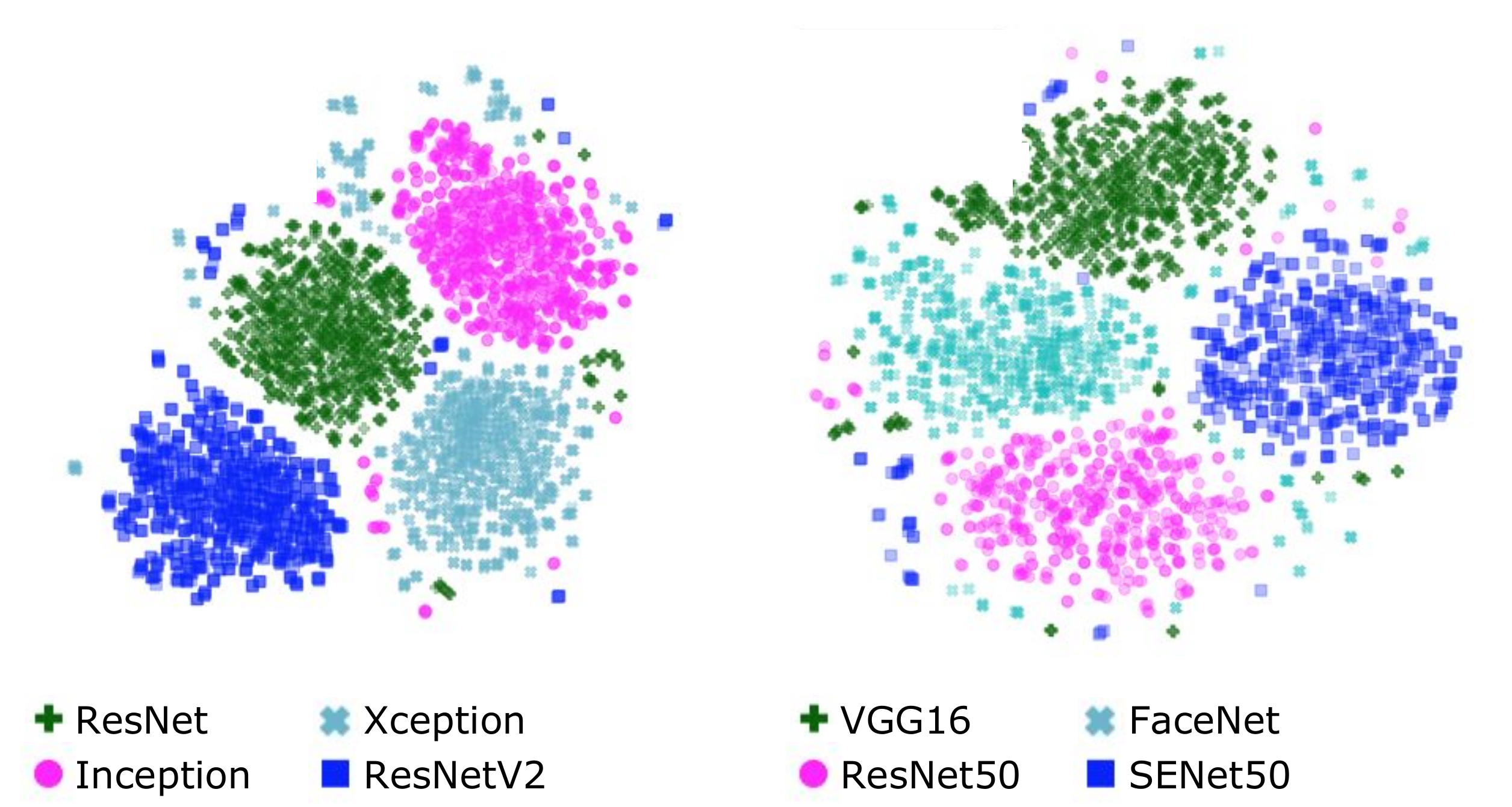}
\caption{Fingerprint embeddings (left) and face embeddings (right) mapped into 2-D plane via t-Distributed Stochastic Neighbor Embedding (t-SNE)~\cite{tsne}.  The clusters indicate that (i) embeddings contain discriminative features for the model of origin, (ii) embeddings from same model have high probability to be at a small euclidean distance.
Details on the face and fingerprint DNN models are in Table~\ref{tab:targets}.} 
\label{fig:tsne}
\end{figure}

In the case of large DNNs used in biometrics (e.g., ResNet~\cite{fingernet}, Facenet~\cite{facenet}), it is unrealistic to finely reverse-engineer all parameters of the target model. Therefore, Model Inference works by directly mapping the embedding to a \textit{known} DNN model. The attacker can in fact crawl a collection of (pre-trained) models from public repositories (e.g., \cite{tfhub}~\cite{onnx}~\cite{kerasapp}), with a variety of DNN architectures. Given such range of models, the attacker finds out which one is likely to have produced the observed embedding. We initially assume, based on Section~\ref{sec::advscenario}, that the \textit{initial} DNN model, \textit{i.e.}, prior to any adaptations to the biometric task and data, is within the attacker's crawled collection. We will then relax this assumption in Section~\ref{sec:lim}.

For further convincing evidence that individual embeddings are discriminative for the original model, we visualize face and fingerprint embeddings from four DNNs in Figure~\ref{fig:tsne}. Since the embedding space is high dimensional (e.g., 512-D for embeddings of length 512), we project embeddings onto the 2D plane using t-SNE~\cite{tsne}.
The clusters in Figure~\ref{fig:tsne} indicate that a DNN embedding contains highly discriminative features for the original DNN model.

\subsection{Reconstruction Phase}
To invert the embedding, the attacker trains a \textit{reconstructor}, \textit{i.e.}, an estimator for the original biometric trait minimizing reconstruction loss over a large training set. 
While models for inverting deep representations have been experimented in non-adversarial settings~\cite{dosovitskiy2015inverting}, these exploit full access to the original target model and its real data, and the capacity to adjust the target model to ease reconstruction (through auto-encoder setup~\cite{dosovitskiy2015inverting}). To address the lack of original data, our attacker builds an adversarial dataset (e.g., a collection of faces or fingerprints from public datasets, or crawled from the Web) that {\em approximates} the unknown $p_X(x)$ from Equation~\ref{eq:1}. Since the target model is not available, the attacker uses the output of Model Inference as an {\em approximation} of the real biometric feature extractor.

Our problem can be expressed as finding the weights $\Theta^*$ of a reconstruction model $G$ which satisfies:
\begin{equation}
\begin{aligned}
\Theta^* = \underset{\Theta}{\text{ argmin }} \int L(x, G_\Theta(\hat{\Phi}(x))) p_{\text{adv}}(x) dx,
\end{aligned}	
\end{equation}
where $G_\theta$ is the reconstructor of parameters $\theta$; $L$ is the reconstruction loss; $p_{\text{adv}}$ is the \textit{adversarial} data distribution, \textit{i.e.}, the one based on attacker's own data; and $\hat{\Phi}$ is the DNN predicted in Model Inference.

Concerning $L$, we adopt a combination of input-space L2 and perceptual loss from several layers of $\hat{\Phi}$ (used as proxy for the real target model). The rationale for this choice is that the attacker's goal is two-fold: to replicate the original biometric trait (Section~\ref{sec:recres}), and to impersonate the subject in the target model (Section~\ref{sub:adv-imper}). Being $\hat{\Phi}_l$ the output of $l$-th layer of $\hat{\Phi}$, the loss can be expressed as:
\begin{equation}
\begin{aligned}
L = w_0 \|x-G(\hat{\Phi}(x))\| + 
\sum_{l=1}^{K} w_l \|\hat{\Phi}_l(x) - 
\hat{\Phi}_l ( G(\hat{\Phi}(x)) )\|  
\end{aligned}	
\end{equation}
To set the perceptual loss, we first choose a subset of (5 to 10) layers of $\hat{\Phi}$, setting the weights $w_l$ of all other layers to 0. For the selected layers, the parameters $w_l$ are updated online, during training, so that the expected contribution from each layer is normalized to the loss (as in \cite{chen2017photographic}). We re-iterate the process with different layer subsets, finally choosing the subset that {\em maximizes} the fraction of reconstructions correctly classified by $\hat{\Phi}$.




\section{Experiments: Setup and Methodology}
\label{sec::eva}
In this section, we describe the models, the datasets, and the experimental configurations used in our tests.
\begin{table}[t]
\caption{Biometric recognition models (target models). To measure the recognition performance, we test the face models on \textit{Facescrub}  
fingerprint models on \textit{PolyU} with 50 test identities.\label{tab:targets}}
\centering
\begin{tabular}{| l | l | l | l | l | l | l |}
\hline
\textbf{Model id} & Biometrics & Pre-Training &  Class.Acc.(\%) & TAR @ 1\% FAR \\ \hline
\textbf{ResNet50}~\cite{kerasvggface} & Faces &   VGGFace  & 97.52 & 99.74\\
\textbf{SENet50}~\cite{kerasvggface} & Faces & VGGFace  & 99.15 & 100.0\\
\textbf{VGG16}~\cite{kerasvggface} & Faces & VGGFace  & 96.90 & 98.95\\
\textbf{FaceNet}~\cite{facenetcode}  & Faces & Casia-WebFace & 98.82 & 99.54\\
\textbf{OpenFace}~\cite{openface} & Faces & Casia-WebFace  & 97.03 & 99.71\\ \hline
\textbf{ResNet}~\cite{kerasapp} & Fingerprints & ResNet50  & 98.74 & 95.80\\
\textbf{Inception}~\cite{kerasapp} & Fingerprints & ImageNet  & 97.80 & 94.70\\
\textbf{Xception}~\cite{kerasapp} & Fingerprints & ImageNet & 97.18 & 96.95\\
\textbf{DenseNet}~\cite{kerasapp} & Fingerprints & ImageNet & 97.33 & 97.86\\
\textbf{ResNet-v2}~\cite{kerasapp} & Fingerprints & ImageNet & 98.24 & 93.22\\ \hline
\end{tabular}
\end{table}

\subsection{Datasets}
We perform the evaluation on two most common physiological biometrics~\cite{survey2018}\cite{survey2019}: face and fingerprint. 

\noindent \textit{Face datasets:} ~We rely on three publicly available datasets widely used in face recognition and deep learning literature: labelled datasets \textit{Facescrub}~\cite{facescrub}\footnote{Since many URLs of face images were removed, we augmented Facescrub with \textit{PubFig}~\cite{pubfig}, obtaining 81,609 images for 425 identities} and \textit{VGGFace2}~\cite{vggface2}, and the unlabelled dataset \textit{FlickR-Faces-HQ}.
We also experiment with two different face alignments: one is based on Dlib's~\cite{dlib} facial landmarks, the other on MTCNN~\cite{mtcnn}. 

\noindent \textit{Fingerprint datasets:} ~We use two datasets designed for academic research purposes, both with labelled fingerprints acquired through (contact-based) scanner impressions\footnote{We do not consider rolled fingerprint images acquired with ink-techniques. In particular, the two benchmark databases of rolled fingerprints (NIST SD4~\cite{nist4} and NIST SD14~\cite{nist14}) were not available at the time of the experiments.}:  \textit{PolyU Contact-Based 2D Fingerprints} (PolyU)~\cite{polyU}, which we augmented (x50 factor) through obliterations, central rotations, translations, and cropping~\cite{fingaugment} obtaining 100,800 images for 336 identities; and \textit{Sokoto Coventry Fingerprint Database} (SocoFing)~\cite{sokoto}~\cite{sokoto2}, that we similarly augmented (x10) obtaining 60,000 images for 600 identities.

\subsection{Experiment Setup}

Our experiments require three components. First, we need DNN models performing biometric recognition and acting as \textit{target models} for the attack.
These models generate biometric feature vectors (\textit{embeddings}) from face or fingerprint images. Second, we need adversarial models by which the attacker can perform Model Inference to unveil the DNN model that originated it. Lastly, we need adversarial reconstructors trained to transform embeddings back into faces or fingerprint images. 

\label{sec:setup}

\descr{Biometric recognition models.} 
~As target models, we use 10 pre-trained models (5 for faces, 5 for fingerprints) based on prominent DNN architectures. In the case of face recognition, we employ models pre-trained on large face datasets.
For fingerprint recognition, we use generic image recognition models pre-trained on ImageNet~\cite{imagenet_cvpr09}, which we additionally train on fingerprint data (as in \cite{fingernet},\cite{michelsanti2017fast}). Table~\ref{tab:targets} presents our target models and their test performance on:    

\noindent \textbullet ~\textit{Identification}, which uses the feature vectors to perform classification over a pre-defined set of identities. 

\noindent \textbullet ~\textit{Verification}, which performs 1:1 matching between two feature vectors (e.g., a new sample and a stored biometric template).
We evaluate Verification by the True Acceptance Rate (TAR) and False Acceptance Rate (FAR), where the FAR is the probability to incorrectly accept a non-authorized person, while the TAR is the probability to correctly accept an authorized person.

\descr{Model Inference setup.} ~We implement Model Inference with an \textit{auxiliary classifier}, \textit{i.e.,} a classifier where (i) the input is one feature-vector (embedding), and (ii) each output class is a different pre-trained model \textit{collected} by the adversary (Table~\ref{tab:targets}). 
To account for the variable embeddings length -- the attacker does not know a-priori which DNN layer is used for embedding extraction -- we train one auxiliary classifier per input size. We consider lengths 128, 512, 2048 for face embeddings and 512, 1024, 2048 for fingerprint.   
We build each classifier as a Artificial Neural Network (ANN) with 5 hidden layers with 512 to 32 hidden units, and train it on a reference set of 10K labelled feature vectors. We min-max normalize all embeddings, so that no assumption is required on the range of their numeric values.

\descr{Reconstruction setup.} ~We train 10 reconstructors, one for each target model in Table~\ref{tab:targets}, whose architecture is inspired on the generative section of DCGAN~\cite{dcgan}. While the DCGAN generator's input is a random-noise vector, our reconstructor takes in input one biometric embedding. More specifically, the reconstructor architecture includes an initial fully-connected layer whose input size matches the feature vector length, followed by \textit{3 to 4} up-sampling blocks performing transpose convolution (depending on the input embedding dimension)\footnote{The detailed architectures are presented in Appendix~\ref{app:rec}}. To train the face reconstructors, 
we merge portions of Facescrub (30K images), VGGFace2 (90K), and FFHQ datasets (70K). Before training, we align all images through MTCNN~\cite{mtcnn} and re-scale them to the standard size of 64x64. To train fingerprint reconstruction, we use 90K 64x64 re-scaled images from PolyU and SocoFing datasets. In both cases, we ensure there is \textit{no overlap between attacker's data and target model's (training or test) data.} 
We implement each reconstruction model using Keras and train it for 200 epochs with batch size 16 and Adagrad optimizer. 

\subsection{Attack Conditions}
\label{sec:attackcond}


\descr{Adaptations of the feature extractor.} ~We test the attack under two conditions. The first, namely \textit{No-adapt.}, corresponds to scenario 1 of Section~\ref{sec::advscenario}: the biometric recognition system adopts a pre-trained DNN model that is used \textit{as is} to extract biometric features. The second condition, namely \textit{Fine-Tuning} (FT), implements scenario-2 of Section~\ref{sec::advscenario}: the initial pre-trained DNN is adapted towards the biometric recognition task via additional training. We test 5 FT configurations, based on the size of the fine-tuning dataset: [5K,10K] for FT-1; [10K,15K] for FT-2; [20K,25K] for FT-3; [30K,35K] for FT-4; [55K,60K] for FT-5.\footnote{A detailed example of FT configurations for ResNet50 model can be found in Appendix~\ref{app:fttl}}
\begin{table}[t]
\caption{Model Inference accuracy results obtained for different feature vector lengths, and under different conditions on attacker's data and target-model adaptations in Section~\ref{sec:attackcond}. Random/chance accuracy is 25\% in all the cases.}
\label{tab:miresults}
\centering
\begin{tabular}{| llllllll |}
 \multicolumn{8}{c}{\textbf{Model Inference -- face recognition models:}}  \\ [1mm]\hline
 &  & \multicolumn{4}{l}{\textbf{Accuracy (\%)}} & & \\ 
Len. & Attack.Data & No-Adapt & FT-1 & FT-2 & FT-3 & FT-4 & FT-5\\ \hline
128 & Same-Ident. & 93.33 & 92.44 & 92.37 & 84.44 & 83.39 & 66.32\\
128 & Same-PreProc. & 90.01 & 89.54 & 89.31 & 79.24 & 86.64 & 61.06\\
128 & Diff-PreProc. & 78.66 & 77.62 & 72.19 & 66.75 & 65.96 & 49.97\\ 
 \hline
512 & Same-Ident. & 95.55 & 93.96 & 92.91 & 86.42 & 85.21 & 68.31\\
512 & Same-PreProc. & 92.11 & 91.24 & 90.04 & 86.17 & 84.97 & 65.23\\
512 &  Diff-PreProc. & 81.60 & 80.75 & 80.12 & 71.11 & 69.81 & 48.21\\ 
\hline
2048 & Same-Ident. & 99.78 & 99.17 & 98.22 & 87.09 & 85.60 & 69.98\\
2048 & Same-PreProc. & 99.08 & 98.55 & 97.73 & 82.15 & 83.41 & 67.22\\
2048 & Diff-PreProc. & 87.22 & 85.98 & 85.16 & 73.48 & 70.04 & 50.02\\ \hline
 \multicolumn{8}{c}{} \\
 \multicolumn{8}{c}{\textbf{Model Inference -- fingerprint recognition models:}}  \\ [1mm]\hline
 &  & \multicolumn{4}{l}{\textbf{Accuracy (\%)}} & & \\ 
Len. & Attack.Data & No-Adapt & FT-1 & FT-2 & FT-3 & FT-4 & FT-5\\ \hline
512 & Same-Ident. & 97.32 & 96.68 & 95.61 & 93.33 & 89.16 & 72.67\\
512 & Same-PreProc. & 96.09 & 94.87 & 93.24 & 91.08 & 85.50 & 66.43\\
512 & Diff-PreProc. & 91.88 & 90.62 & 88.12 & 86.94 & 82.45 & 60.12\\ 

 \hline
1024 & Same-Ident. & 97.84 & 97.03 & 96.92 & 93.81 & 86.61 & 71.57\\
1024 & Same-PreProc. & 96.24 & 95.46 & 94.84 & 92.93 & 84.92 & 68.59\\
1024 &  Diff-PreProc. & 91.95 & 90.41 & 89.07 & 87.26 & 83.81 & 60.35\\ 

\hline
2048 & Same-Ident.  & 99.64 & 99.27 & 98.84 & 95.11 & 85.90 & 70.69\\
2048 & Same-PreProc.  & 98.28 & 98.15 & 97.52 & 92.24 & 87.09 & 69,32\\
2048 &  Diff-PreProc. & 92.45 & 90.77 & 89.63 & 87.82 & 83.73 & 61.44\\ \hline
\end{tabular}
\end{table}

\section{Experimental Results}
\label{sec:results}
Our evaluation focuses on three main attacker's capabilities: 
the capability to infer the DNN model used as biometric feature extractor from one observation of one embedding;
the capability to recover the original biometric input and obtain a reconstruction that is similar``close" to the original; 
the capability to gain authentication onto the original biometric system using the reconstructed biometric trait.


\subsection{Model Inference Results}

\label{sec:miresults}
We start by assessing the attacker's capability to infer the target model used by the biometric recognition system. Since Model Inference relies on auxiliary classifiers that associate one biometric embedding to the model that possibly generated it, we measure its performance in terms of classification accuracy.

As described in Section~\ref{sec:setup}, instead of running one global classifier, we implement Model Inference through multiple auxiliary classifiers, each accounting for a specific embedding size. From the models in Table~\ref{tab:targets}, we can extract embeddings of length [128, 512, 2048] for faces, [512, 1024, 2048] for fingerprints. Therefore, we obtain 6 auxiliary classifiers, one for each biometric modality (face/fingerprint) and embedding size. Each classifier has 4 classes (\textit{i.e.,} four feature-extractor labels). This means 24 total different feature extractors (classes), that we obtained out of the 10 models in Table~\ref{tab:targets} by exploiting the embedding extraction from intermediate layers\footnote{This condition does not penalize the attacker: if the attacker can infer the pre-trained model, she can also similarly explore the use of different model layers. We include further details on the embedding extraction layers in Appendix~\ref{appendix:mi}.}. 
Before training or testing each classifiers, we also min-max normalize all embeddings. This way, we need no assumption on the range of numeric values in the embeddings, which depends, for instance, on the rgb scale of the original images (0-1/0-255).

\begin{table}[t]
\caption{Model Inference accuracy results on DNN models collected from Tensorflow-Hub~\cite{tfhub} library. \label{tab:tfhub}}
\centering
\begin{tabular}{llll}
\toprule
\textbf{Feature-vector len.} & \textbf{N. Classes} & \textbf{Accuracy} & \textbf{Random/Chance}\\ 
\midrule
2048 & 39 & 97.62\% & 2.6\% \\
1024 & 24 & 98.21\% & 4.2\% \\
512 & 31 & 96.04\% & 3.2\% \\
128 & 11 & 98.93\% & 9.0\% \\
\bottomrule
\end{tabular}
\end{table}

Table~\ref{tab:miresults} reports the classification results obtained for different embedding lengths and attack conditions on attacker's data and target model adaptations. 
Model Inference achieves up to 99\% classification accuracy for both face and fingerprint embeddings. While longer embeddings are generally more informative, Model Inference is still effective (up to 93\% and 97\% accuracy) for representations of size 128 for faces and 512 for fingerprints\footnote{To our knowledge, 128 is the smallest biometric embedding length, corresponding to the case of Facenet~\cite{facenet} or OpenFace~\cite{openface}.}. 
\descr{Impact of adaptations on the feature extractor.} ~As far as Fine Tuning (FT) is concerned, it should be harder in theory for the attacker to unveil the original pre-trained model, as FT modifies the original DNN's weights. However, Table~\ref{tab:miresults} shows that Model Inference is still effective under FT. Average accuracy is  $\ge85\%$ on fingerprint embeddings and  $\ge79\%$ on face embeddings for all setups up to \textit{FT-4} (\textit{i.e.}, additional target-model training on 30K-35K new faces-fingerprints). In other words, even under fine tuning, the embeddings maintain a clear \textit{footprint} of the initial pre-trained DNN model adopted by the biometric recognition system, that the attacker exploits.      

\descr{Impact of attacker's data.} ~By varying the conditions on the attacker's data, we observe a clear accuracy reduction only when the samples in the attacker's dataset are differently pre-processed compared to the test data, roughly $-5\%$ for fingerprint embeddings and $-10$/$-15\%$ for face embeddings. However, it is worth noting that even under \textit{Diff-PreProcessing}, Model Inference produces correct predictions for at least 78.66\% for face and 91.88\% for fingerprint models.

\descr{Experiments on a large model set.} ~Due to the large availability of pre-trained models online (e.g., \cite{tfhub}~\cite{onnx}~\cite{kerasapp}), it is fair to assume that Model Inference would in practice operate on larger pools of models. To experiment with this condition, we have deployed Model Inference on all the relevant models available on \texttt{Tensorflow-HUB}~\cite{tfhub}, one of the most extensive library or re-usable neural networks. Specifically, out of the total 466 DNNs found on the platform\footnote{The number of models on TensorFlow Hub is ever increasing; these numbers refer to September 2020.}, we select the 105 models with the tag ``image feature vector'', which can be adapted for usage as biometric feature extractors. This set includes several instances of popular model architectures, in particular ResNet (16 models) and EfficientNet (21 models). Since all the models are pre-trained as generic ImageNet classifiers, we additionally train them on one of our datasets (Facescrub) as done for the models in Table~\ref{tab:targets}. 

The results reported in Table~\ref{tab:tfhub} show that Model Inference achieves high classification accuracy on this larger model set, consistently above 96\% for all feature vector lengths, despite the increasing number of classes (up to 39) compared to Table~\ref{tab:miresults}. Notably, we found that the fraction of misclassified embeddings is minimal even if the attacker's collection includes many models with the same DNN architecture. In particular, we obtained only 2.88\% misclassifications in the case of ResNet models and 3.92\% for EfficientNet ones.


\subsection{Reconstruction Results}
\label{sec:recres}

To evaluate the quality of biometric reconstructions, we measure the DSSIM (Structural Dis-Similarity Index)~\cite{dssim1}~\cite{dssim2} between the original image and its reconstruction\footnote{The DSSIM has recently gained popularity as a measure of user-perceived image distortion~\cite{shan2019using}~\cite{shan2020fawkes}.}, and the \textit{Perceptual} loss~\cite{perceptual}, {\it i.e.,} the Euclidean distance between the embeddings of the original and reconstructed image after passing them to the same DNN model (\textit{ResNet50} for faces, \textit{ResNet} for fingerprints). The DSSIM and Perceptual loss values generally fall in different ranges for faces and fingerprints, due to differences in the raw data (e.g., color images for face, black-white for fingerprints) as well as in the model used to extract the Perceptual loss.

\begin{figure}[t]
  \includegraphics[width=\columnwidth]{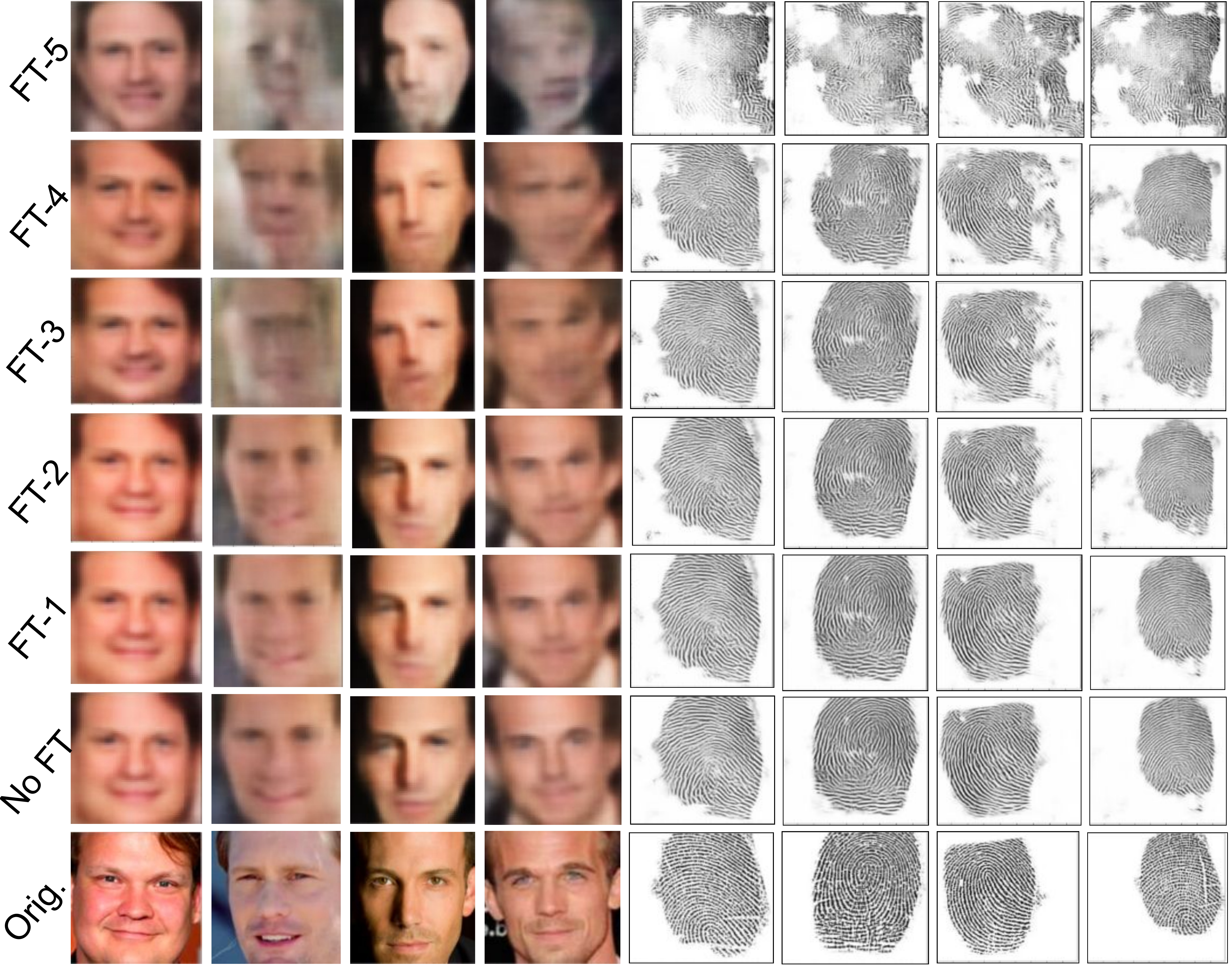}
\caption{Biometric reconstructions obtained when the target model is ``adapted" towards the specific biometric task. We apply fine-tuning (FT) on \textit{ResNet50} face model and \textit{ResNet} fingerprint model. 
}
\label{fig:ftex}
\end{figure}

\begin{table}[t]
\caption{Reconstruction errors (median) when the target model (\textit{ResNet50} face model and \textit{ResNet} fingerprint model) is adapted through fine tuning (FT).  \label{tab:ftres}}
\centering
\begin{tabular}{l|llll}
\textbf{Fine-tuning (FT)} & \multicolumn{2}{c}{\textbf{Face reconstr.}} & \multicolumn{2}{c}{\textbf{Fingerprint reconstr.}} \\ 
\textbf{of target model} & \textbf{DSSIM} & \textbf{Perc.Loss} & \textbf{DSSIM} & \textbf{Perc.Loss}\\ 
\hline
\textit{No FT} & 0.68 & 0.31 & 0.88 & 9.80 \\
\textit{FT-1} & 0.70 & 0.32 & 0.93 & 10.19 \\
\textit{FT-2} & 0.71 & 0.33 & 1.08 & 10.82 \\
\textit{FT-3} & 1.61 & 0.37 & 1.31 & 11.62 \\
\textit{FT-4} & 1.80 & 0.38 & 1.70 & 12.33 \\
\textit{FT-5} & 2.32 & 0.40 & 2.80 & 14.40 \\
\midrule
\end{tabular}
\end{table}

\begin{figure*}[ht]
\subfloat[Face, visual loss ]{\includegraphics[width=0.249\linewidth]{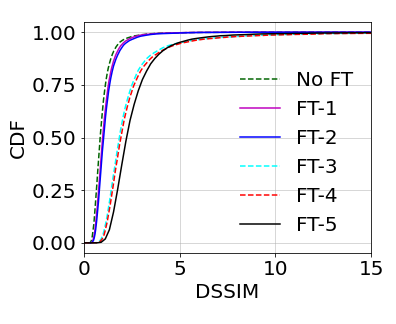}}
\subfloat[Face, perceptual loss ]{\includegraphics[width=0.249\linewidth]{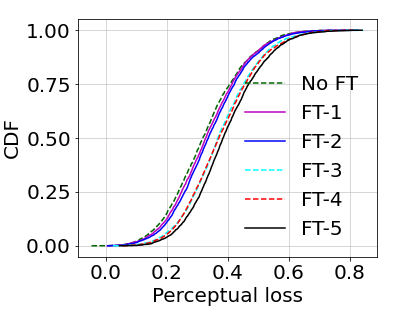}}
\subfloat[Fingerprint, visual loss]{\includegraphics[width=0.249\linewidth]{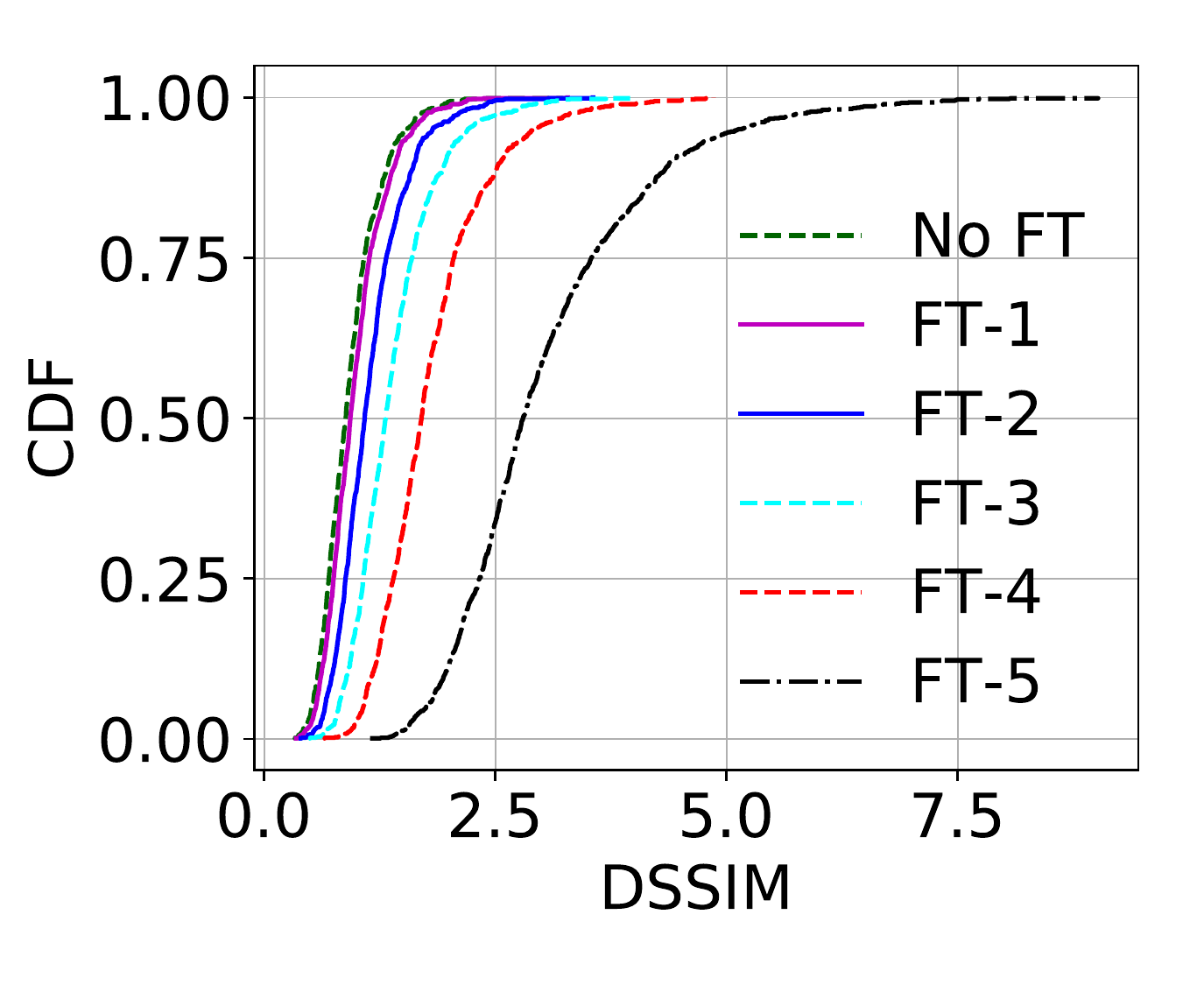}}
\subfloat[Fingerprint, perceptual loss]{\includegraphics[width=0.249\linewidth]{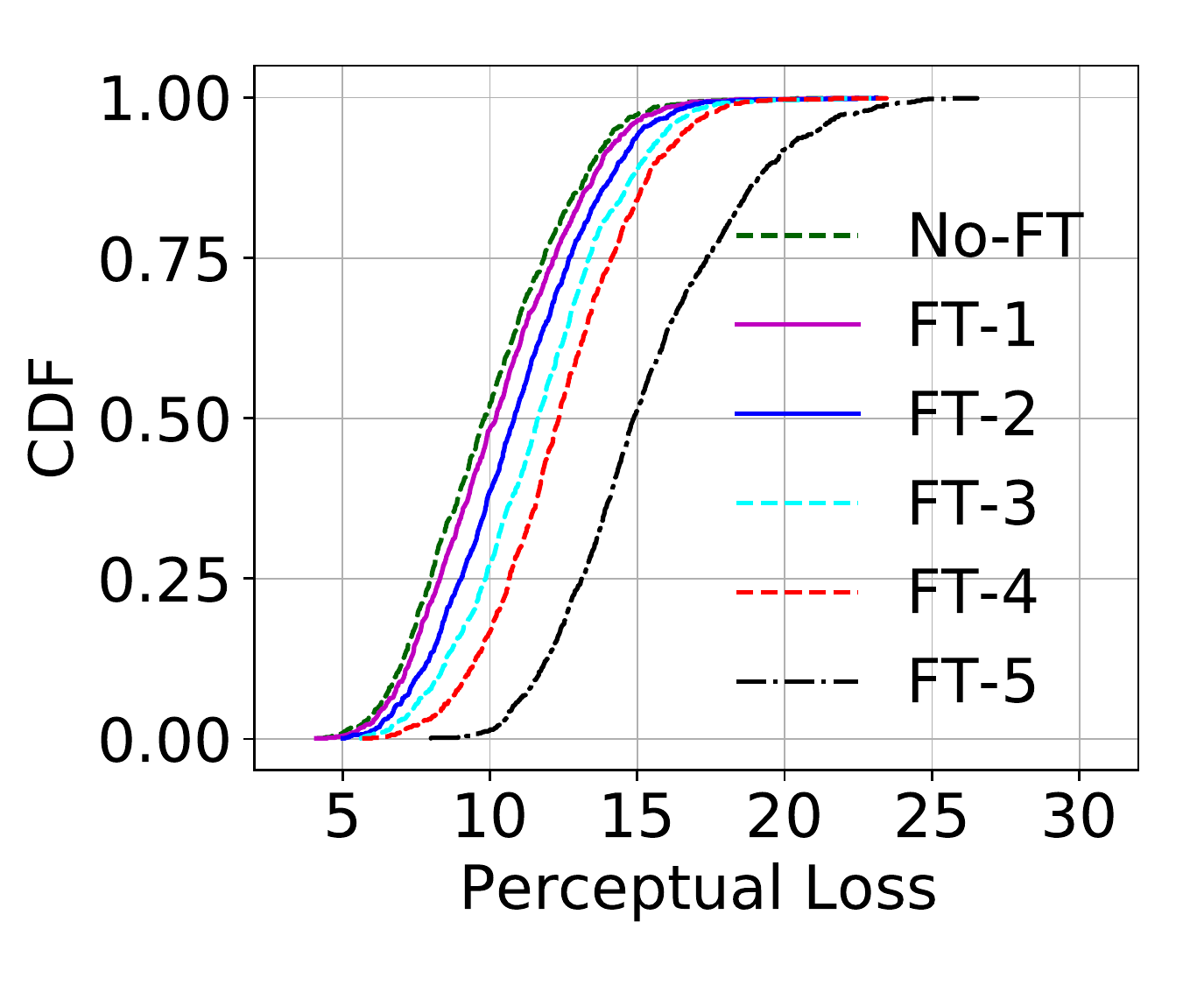}}
\caption{Effect of adaptations of the target model on the reconstruction errors (DSSIM and \textit{Perceptual loss}). We have progressively modified \textit{ResNet50} face model and \textit{ResNet} fingerprint model through \textit{fine tuning} (FT).} 
\label{fig:ftres}
\end{figure*}

\descr{Impact of adaptations on the feature extractor.} ~The fine-tuning (FT) of the target model on the biometric task introduces a misalignment between the attacker's feature extractor (obtained in Model Inference) and the actual target model. The question is whether this prevents the attacker from recovering the original biometric trait. 
In Figure~\ref{fig:ftex}, while observing a progressive degradation of reconstructions, we note that face semantics only marginally deteriorate for FT-1 and FT-2 setups; similarly, for FT 1-3 the attack still recovers most of the fingerprint details. In addition to qualitative examples, we present in Figure~\ref{fig:ftres} the empirical CDF of DSSIM and Perceptual loss for all FT configurations, with median values reported in Table~\ref{tab:ftres}. The results on DSSIM and Perceptual loss confirm the observation, as reconstruction errors increase substantially only under more extreme FT conditions (e.g., $+165\%$ and $+93\%$ median DSSIM in FT-4). 
In practice, the attack can craft quality reconstructions as long as the fine-tuning dataset is {\em moderate}, \textit{i.e.}, up to 20-25K new samples.  

\textit{Remark:} The deterioration of biometric reconstructions in Figure~\ref{fig:ftex} is an effect of the embedding variations caused by fine-tuning. In essence, the attacker trains the reconstructor to invert an embedding $x$ from the specific DNN model \textit{guessed} in Model Inference. In case of adaptations of the target model, the reconstructor tries to invert embeddings that are different from the ones it was trained for (e.g., $x+\delta$ instead of $x$). The more the variation in the embeddings ($\delta$), the more the noise in the reconstructions. More formally, we can link the embedding variation $\delta$ with the reconstruction variation $G(x+\delta)-G(x)$ (deterioration) using the \textit{Lipschitz} constant of the attacker's reconstructor model\footnote{The concept of Lipschitz constant of neural networks has been used in spectral analysis~\cite{szegedy2013intriguing} and normalization~\cite{lipreg}~\cite{lipreg2} of neural networks.}, defined as the constant $L$ such that
\begin{equation}
            \forall x,\delta : \lVert G(x) - G(x+\delta)\rVert \le L\lVert\delta\rVert
\end{equation}
Assuming we can derive $L$, this relation says that the maximum degradation of the reconstructions (compared to \textit{no-adapt} case) is proportional to the embedding variation $\delta$. In Appendix~\ref{app:lip}, we show how to compute $L$ for the biometric reconstruction network. In short, the method builds on (i) the composition of Lipschitz constants of individual network layers; (ii) the derivation of transpose-convolution layer's constant from the analogy with standard convolutional layer.

\begin{figure}[t]
\includegraphics[width=\columnwidth]{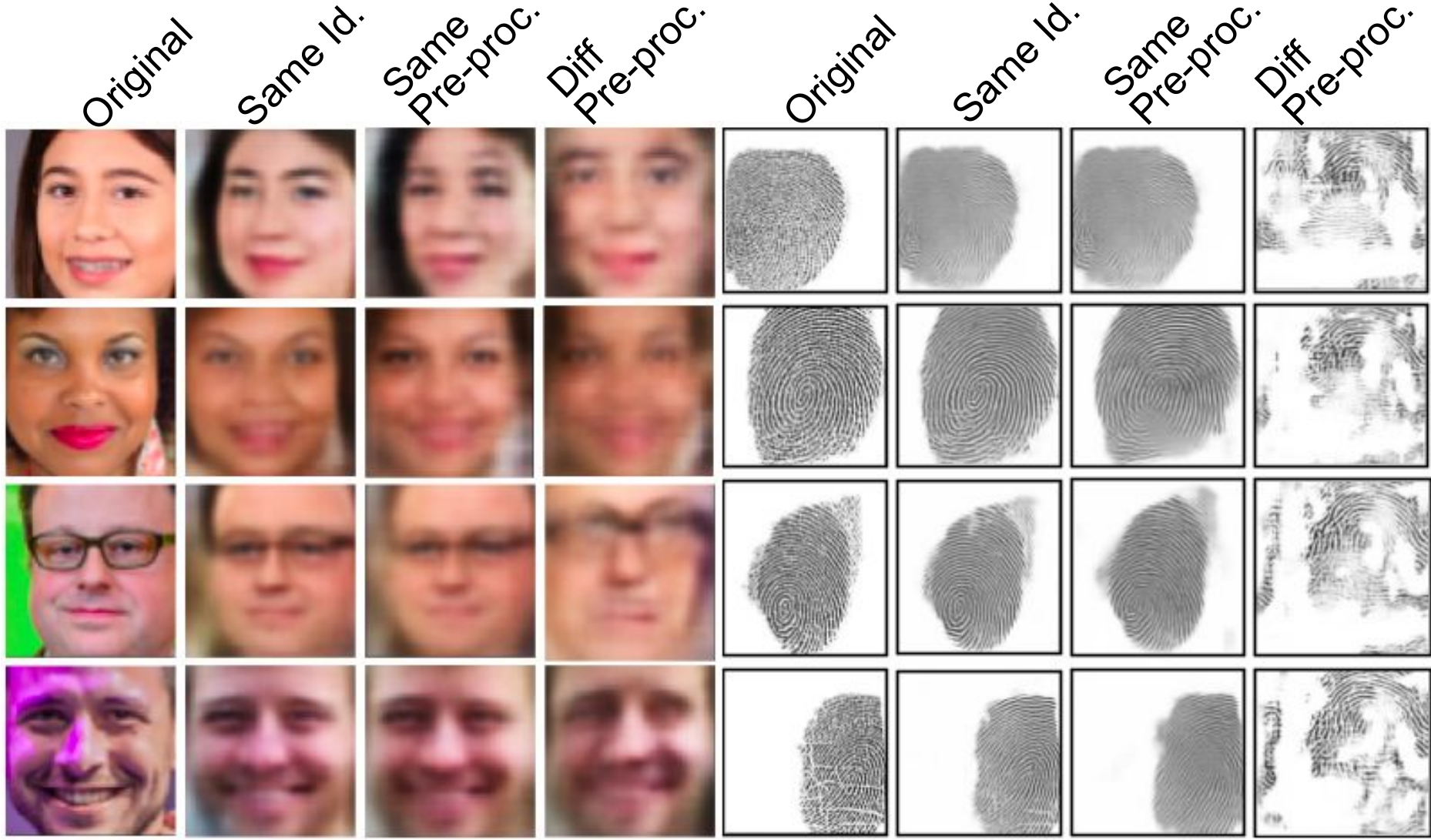} 
\caption{
 Examples of biometric reconstructions obtained under the different conditions on the attacker's data.
} 
\label{fig:dataex}
\end{figure}

\begin{table}[t]
\caption{Reconstruction errors (median) for different conditions on the attacker's data.\label{tab:recdata} }
\centering
\begin{tabular}{l|llll}
\textbf{Attacker's} & \multicolumn{2}{c}{\textbf{Face reconstr.}} & \multicolumn{2}{c}{\textbf{Fingerprint reconstr.}} \\ 
\textbf{data} & \textbf{DSSIM} & \textbf{Perc.Loss} & \textbf{DSSIM} & \textbf{Perc.Loss}\\ 
\hline
\textit{Same-Identities} & 0.83 & 0.23 & 0.38 & 9.85 \\
\textit{Same-PreProc.} & 0.98 & 0.24 & 1.51 & 9.98 \\
\textit{Diff-PreProc.} & 2.55 & 0.28 & 3.39 & 31.95 \\
\midrule
\end{tabular}
\end{table}

\begin{figure*}[ht]
\subfloat[Face, visual loss]{\includegraphics[width=0.245\linewidth]{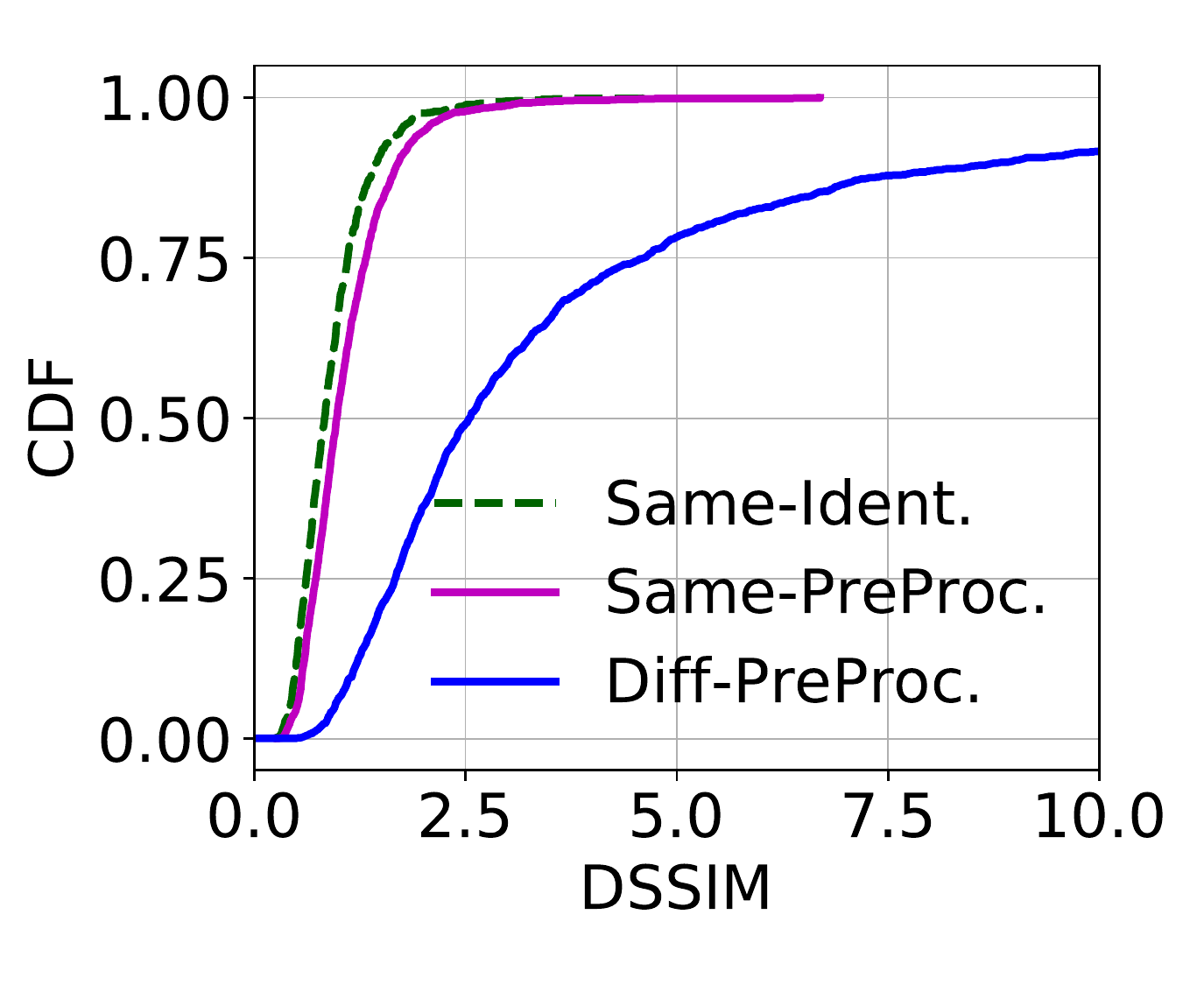}}
\subfloat[Face, perceptual loss]{\includegraphics[width=0.245\linewidth]{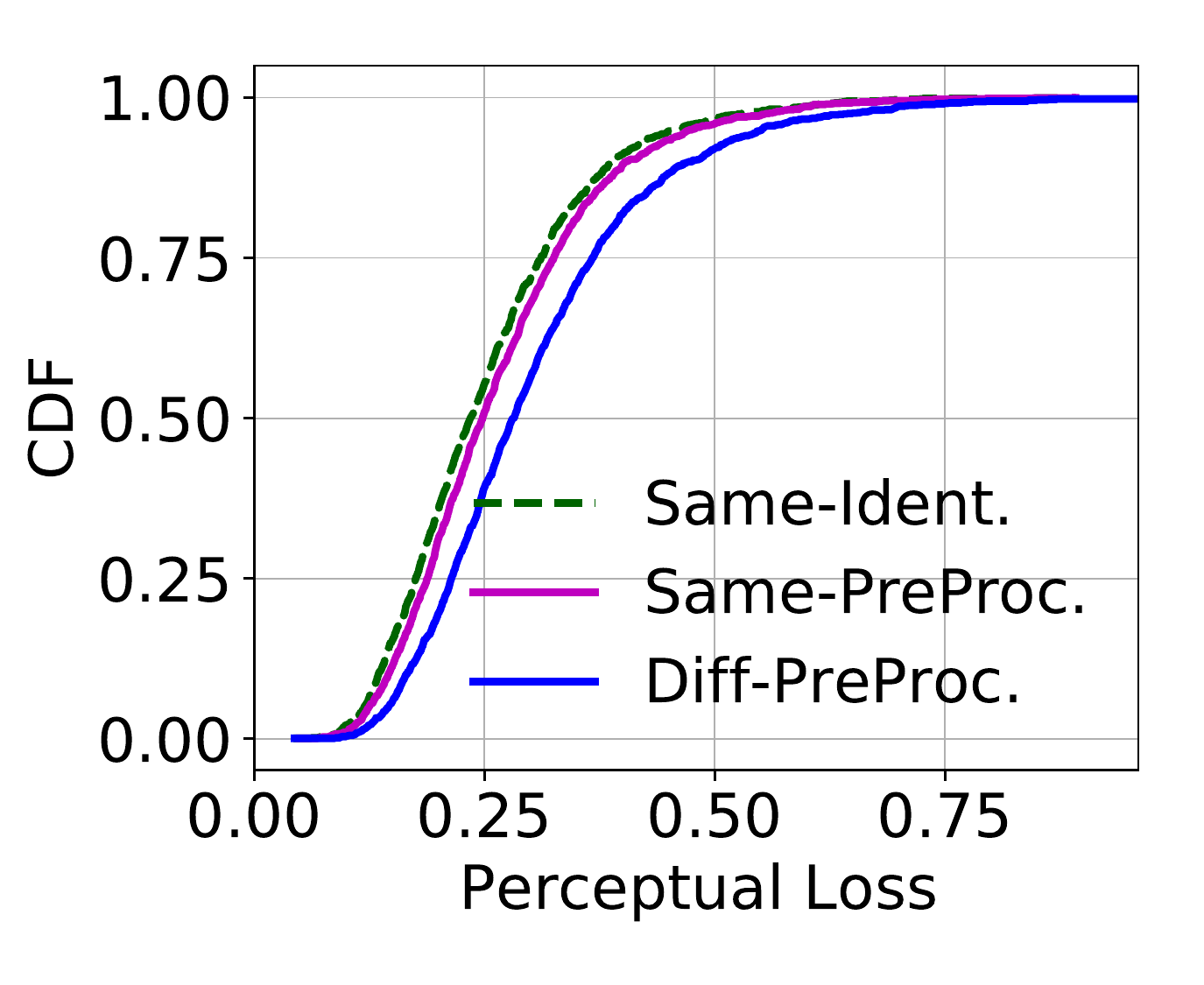}}
\subfloat[Fingerprint, visual loss]{\includegraphics[width=0.245\linewidth]{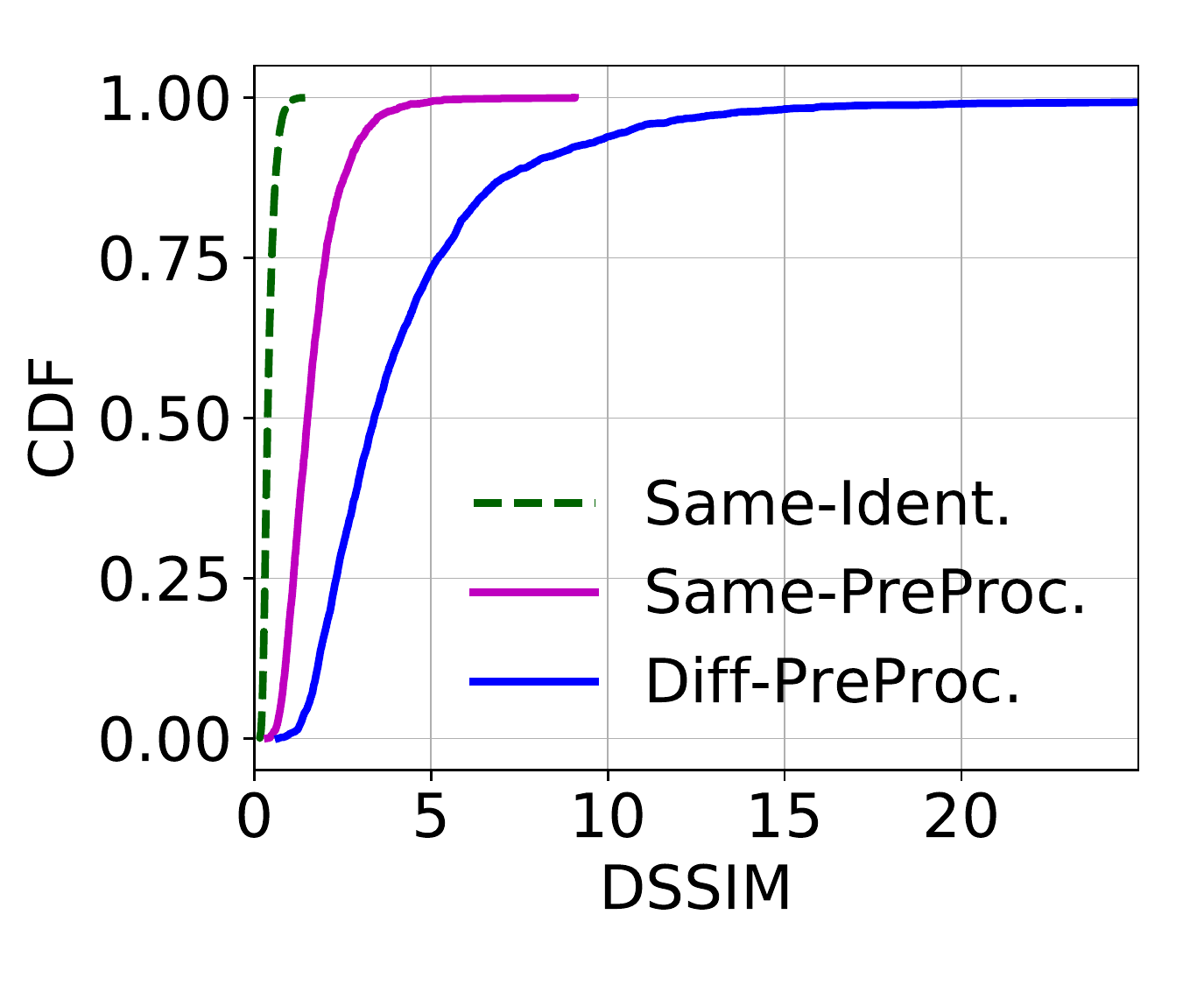}}
\subfloat[Fingerprint, perceptual loss]{\includegraphics[width=0.245\linewidth]{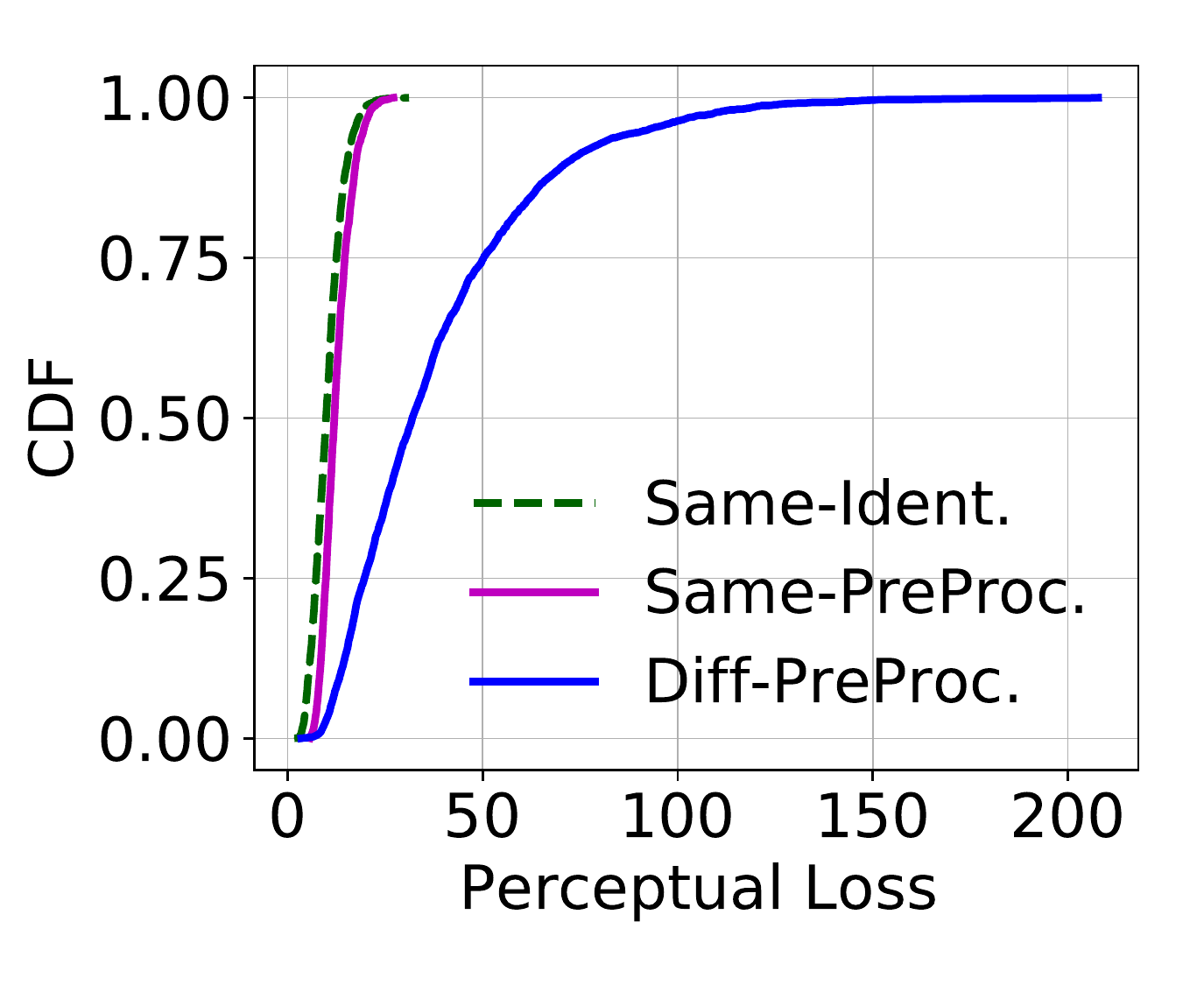}}
\caption{Empirical CDF of the face and fingerprint reconstruction errors evaluated for different conditions on the attacker's data.} 
\label{fig:datares}
\end{figure*}

\begin{figure*}[ht]
\subfloat[Face, visual loss]{\includegraphics[width=0.249\linewidth]{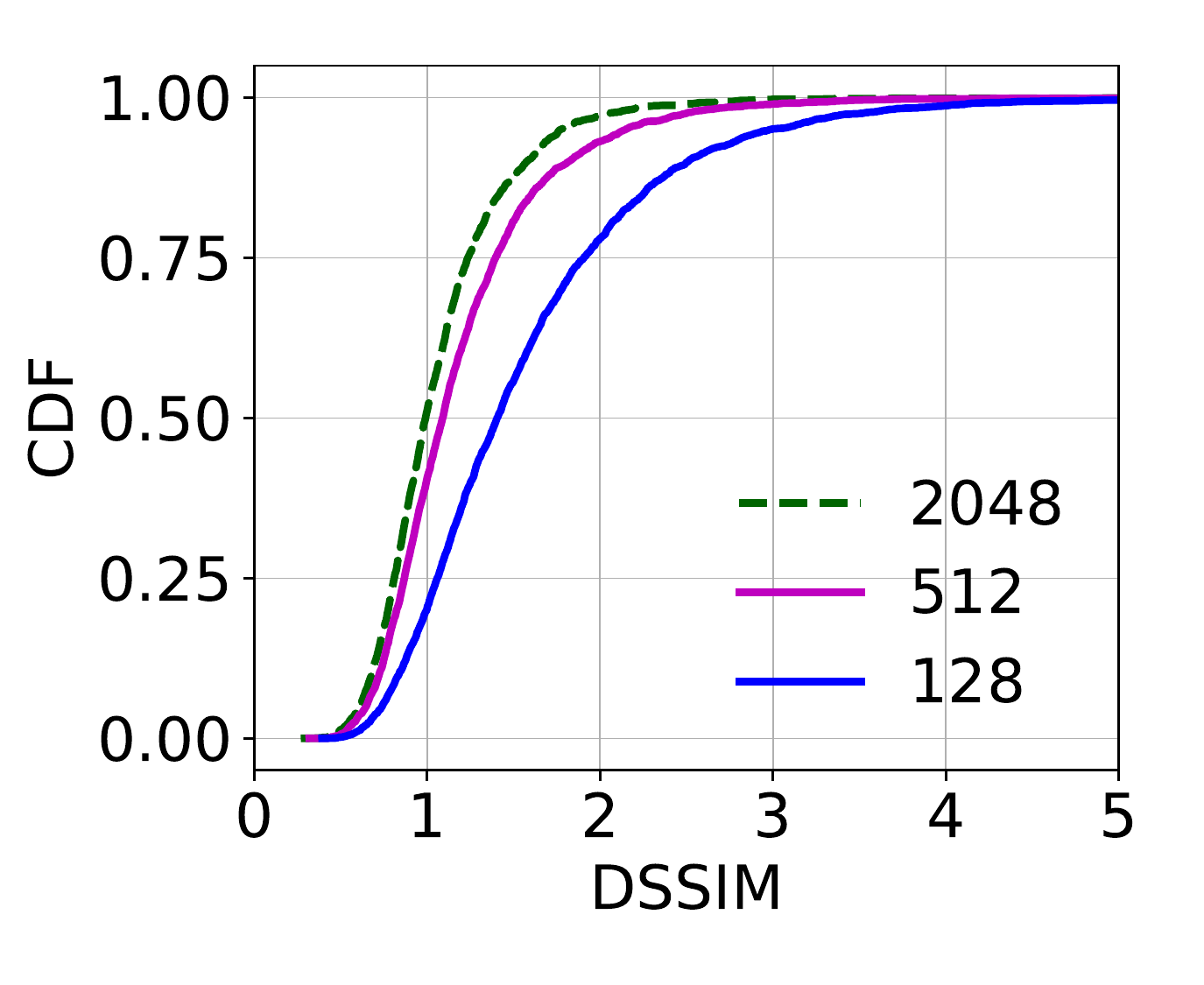}}
\subfloat[Face, perceptual loss]{\includegraphics[width=0.249\linewidth]{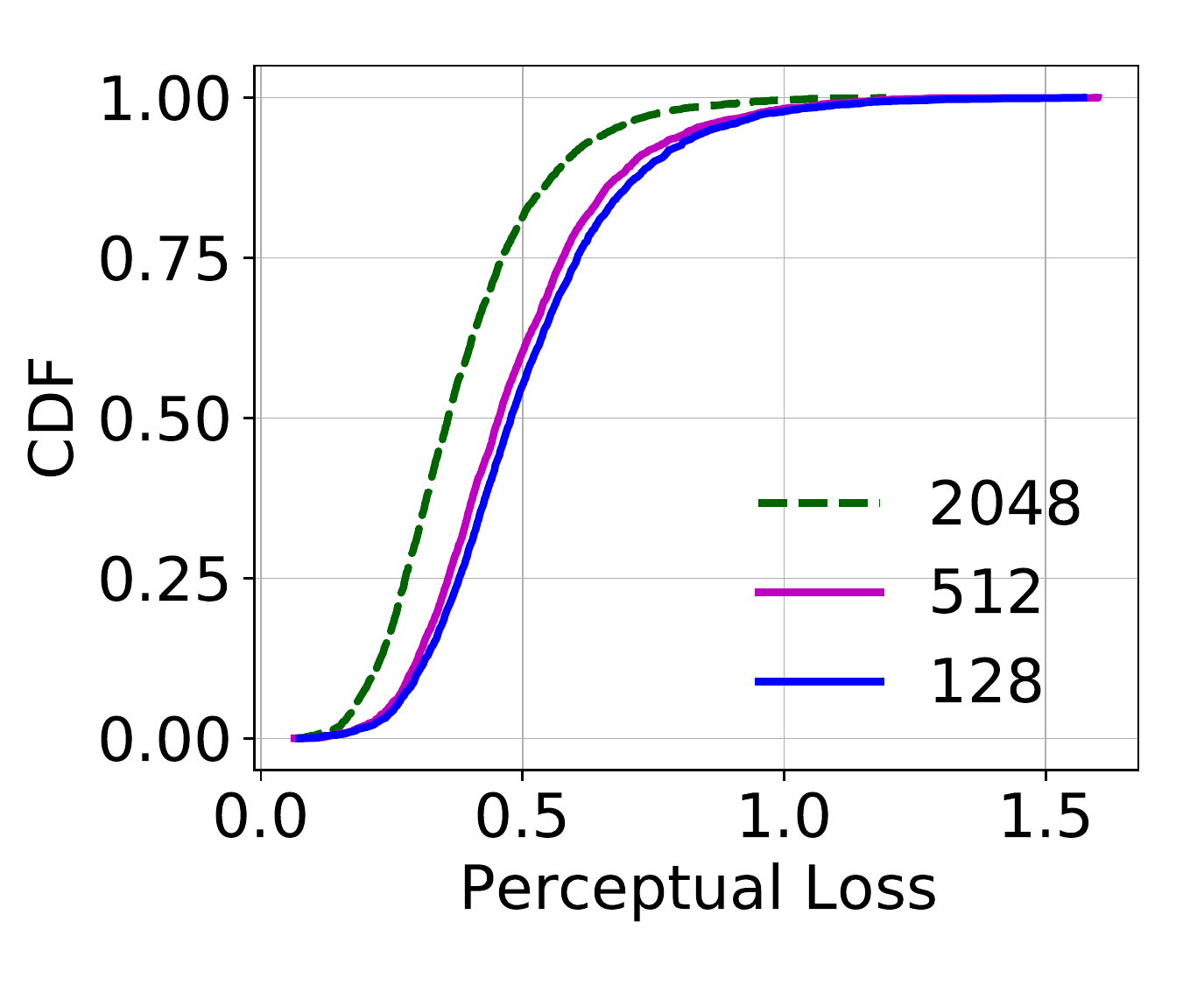}}
\subfloat[Fingerprint, visual loss]{\includegraphics[width=0.249\linewidth]{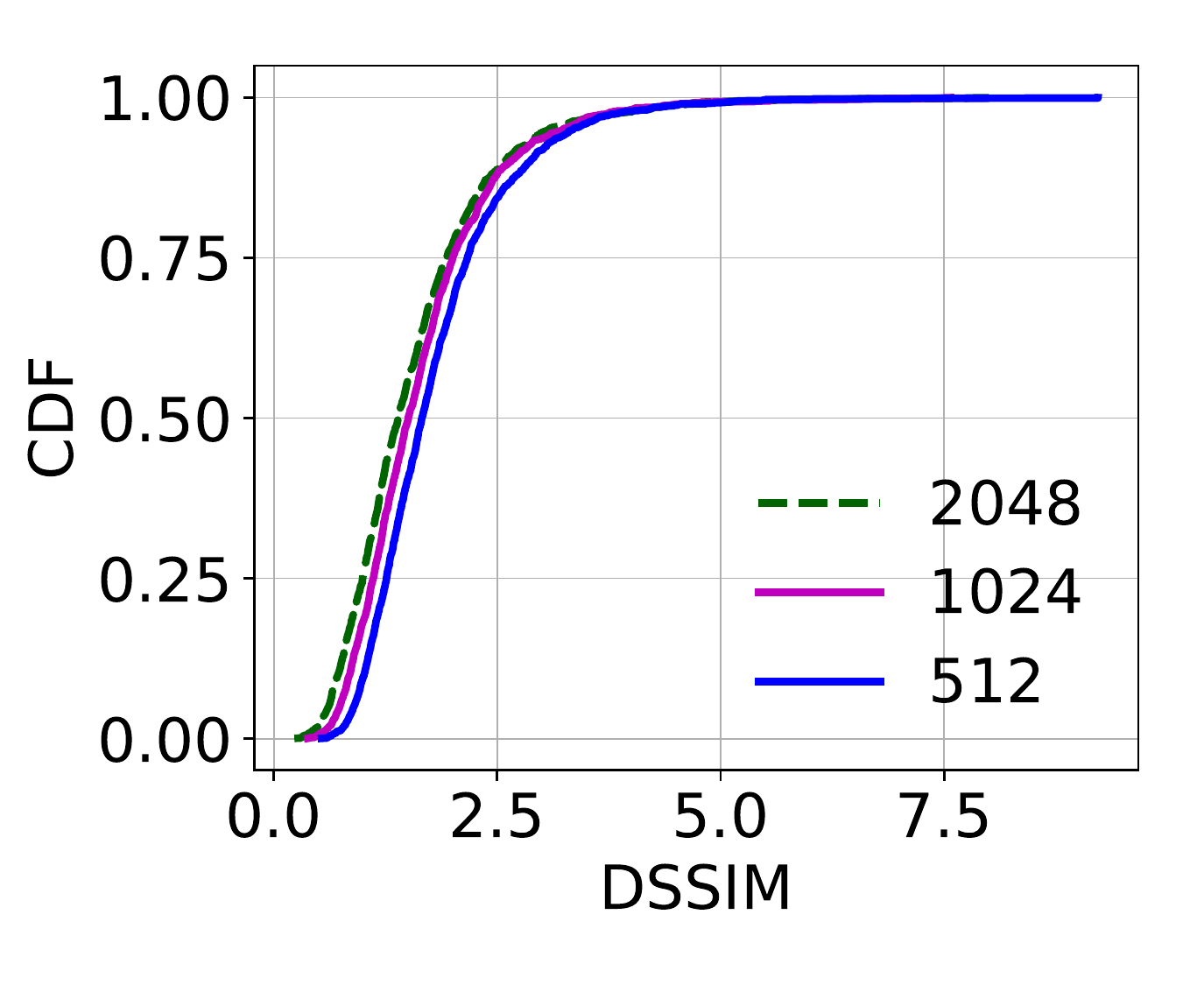}}
\subfloat[Fingerprint, perceptual loss]{\includegraphics[width=0.249\linewidth]{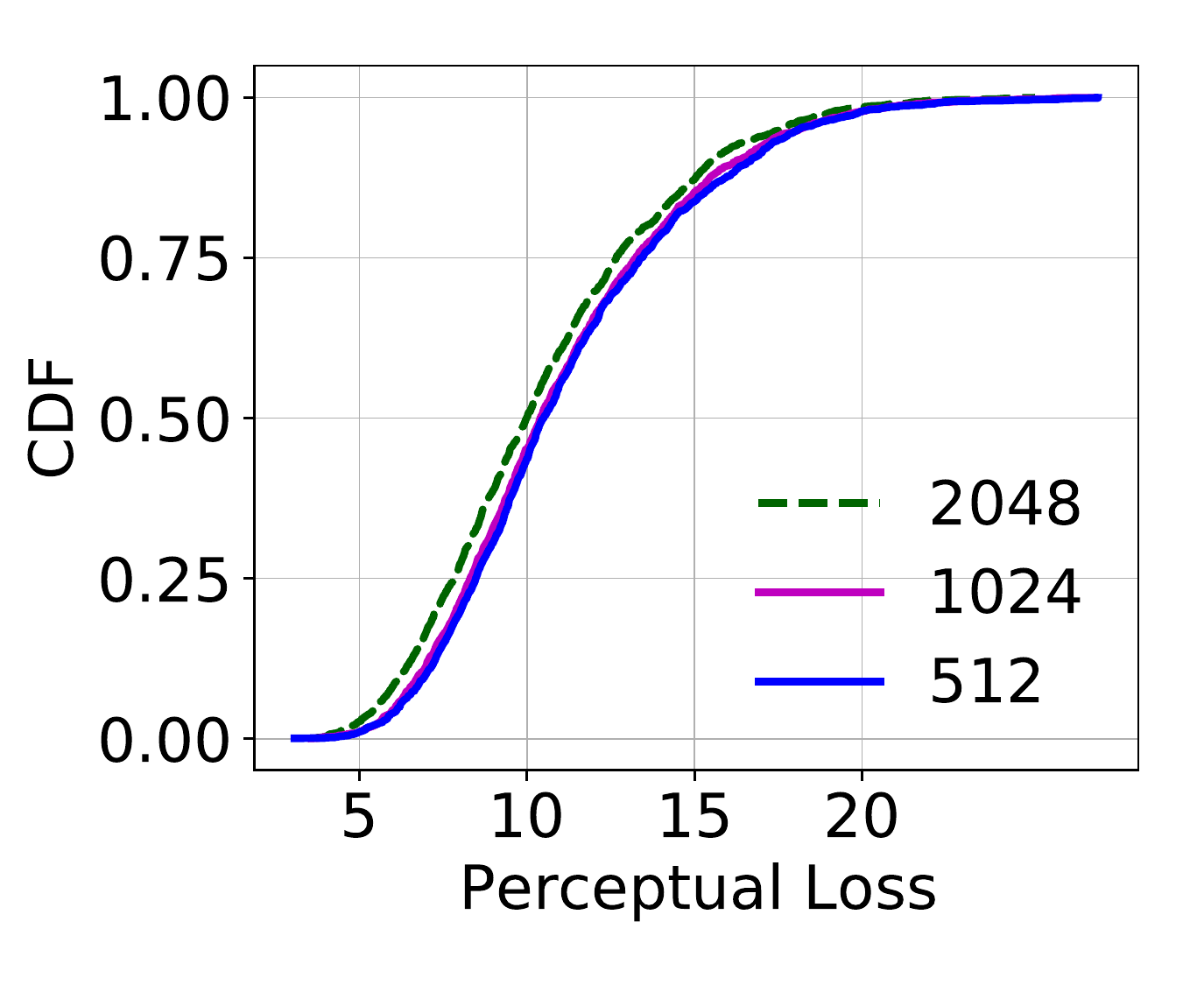}}
\caption{Empirical CDF of the reconstruction errors (DSSIM and \textit{Perceptual loss}) for different feature-vector (embedding) lengths.
\label{fig:emblen}
} 
\end{figure*}

\descr{Impact of attacker's data.}
~Figure~\ref{fig:dataex} reports examples of reconstructions obtained under Same-Identities, Same-PreProcessing, and Diff-PreProcessing conditions. We observe that with no access to any sample for the identities from the real dataset, the attack only incurs limited reconstruction degradations. The quality loss is more tangible if attacker's data is differently pre-processed: limited face semantics are recovered, while fingerprint reconstruction becomes impractical. Figure~\ref{fig:datares} and Table~\ref{tab:recdata}, reporting the empirical CDF of the reconstruction errors and their median values, confirm our findings. In terms of DSSIM, the median fingerprint reconstruction error for \textit{Diff-PreProcessing} is 9 times more than for \textit{Same-Identities} (3.39 vs 0.38). We conjecture that the abrupt degradation of fingerprints reconstructions is due to the limited scale of the fingerprint datasets used to train the target model. This causes fingerprint embeddings to be less \textit{invariant} (compared to face ones) with respect to different data pre-processing, with the result that fingerprint embeddings are effective (for reconstruction) only when the attacker has access to similarly pre-processed data.

\descr{Impact of the DNN.} Lastly, we investigate the impact of the DNN model on the embedding inversion. Table~\ref{tab:recmodels} reports the reconstruction errors for different target models from Table~\ref{tab:targets}. We observe that, while for fingerprints the initial DNN model has a marginal effect on reconstructions (due to fingerprint models being trained on the same data), face embeddings yields more variable reconstruction errors even for ``similarly'' trained models such as ResNet50 (DSSIM=0.97) and VGG16 (DSSIM=1.27).

The cases of DenseNet and VGG16 suggest a relation between feature vector length and the reconstruction quality. To further analyze this effect, we use the 24 feature extractors obtained in Section~\ref{sec:miresults} by exploiting additional feature-extraction layers offered by models in Table~\ref{tab:targets}. For this purpose, we trained additional 14 reconstruction networks fed with intermediate-layer embeddings. This allows to experiment with embedding length 128, 512, 2048 for faces, 512, 1024, 20148 for fingerprints -- further details on the additional feature-extractors are in Appendix~\ref{appendix:mi}. Figure~\ref{fig:emblen} shows the empirical CDF of DSSIM and Perceptual loss after aggregating the results based on embedding length. We observe that increasing embedding lengths consistently provide smaller reconstruction  errors. Differences are, however, less pronounced for fingerprints (e.g., $\le12\%$ on median DSSIM) than for faces ($\le45\%$): this is not surprising, as all fingerprint models were trained on the same data. 
While longer embeddings generally yield better reconstructions, one exception in Table~\ref{tab:recmodels} is the case of Facenet embeddings, which yield the lowest DSSIM despite their minimal length (128). We conjecture the reason behind this is that Facenet was expressively designed to produce efficient representations via \textit{triplet-loss} training~\cite{facenet}, which makes Facenet embeddings particularly ``informative" for the face reconstructions.


\begin{table}[ht]
\caption{Average biometric reconstruction errors for different target DNN models.
\label{tab:recmodels}}
\centering
\begin{tabular}{llll|ll}
\textbf{Model} & \textbf{\#Layers} & \textbf{\#Params} & \textbf{Emb.Len.} & \textbf{DSSIM} & \textbf{Perc.Loss}\\ 
\hline
\multicolumn{3}{l}{\textit{\textbf{Face recognition:}}} & & & \\ 
ResNet50 & 175 & 23M & 2048 & 0.97 & 0.43 \\
VGG16 & 20 & 14M & 512 & 1.27 & 0.57 \\
SeNet50 & 287 & 26M & 2048 & 0.98 & 0.44  \\
OpenFace & 157 & 3.7M & 128 & 1.44 & 0.67\\
Facenet & 338 & 23M & 128 & 0.75 & 0.39 \\ 
\multicolumn{3}{l}{\textit{\textbf{Fingerprint recognition:}}} & & & \\ 
ResNet & 176 & 25M & 2048 & 1.15 & 10.56 \\
Inception & 159 & 24M & 2048 & 1.18 & 11.03 \\
Xception  & 126 & 23M & 2048 & 1.15 & 11.24 \\
ResNetV2  & 191 & 26M & 2048 & 1.18 & 10.90 \\ 
DenseNet & 121 & 8M & 1024 & 1.24 & 11.52 \\ 
\midrule
\end{tabular}
\end{table}

\begin{table*}[h]
\caption{Biometric identification and verification results using reconstructions, for different target models and for the different conditions on the attacker's data in Section~\ref{sec:attackcond}. Random/chance identification accuracy is 0.3\% for both biometric modalities.}
\resizebox{\linewidth}{!}{%
\begin{tabular}{lllll | lllll}
\toprule
\multicolumn{5}{c}{\textbf{Face recognition}} & \multicolumn{5}{c}{\textbf{Fingerprint recognition}} \\ \midrule
Target & Attacker's & Identification &  \multicolumn{2}{c}{TAR @ 1\% FAR} & Target & Attacker's & Identification & \multicolumn{2}{c}{TAR @ 1\% FAR} \\ 
model & data & accuracy [\%] & Same-Image & Same-Subject & model & data & accuracy [\%] & Same-Image & Same-Subject \\ \midrule

ResNet50 & Same-Ident. & 28.1 & 45.4 & 41.7 & ResNet & Same-Ident. & 41.8 & 73 & 55.7  \\
SeNet50 & Same-Ident. & 73.3 & 92.0 & 89.0 & Inception &  Same-Ident. & 25.15 & 59.1 & 44.4  \\
VGG16  & Same-Ident. & 32.4 & 87.9 & 75.7 & Xception  & Same-Ident. & 26.3 & 64.4 & 46.5  \\
Facenet & Same-Ident. & 41.6 & 18.0 & 9.5 & ResNetV2 &  Same-Ident. & 37.6 & 68.4 & 51.2 \\
OpenFace  & Same-Ident. & 51.4 & 35.3 & 27.4 & DenseNet  &  Same-Ident. & 26.9 & 67.3 & 48.4 \\ \midrule

Resnet50 & Same-PreProc. & 27.2 & 32.0 & 29.8 & ResNet & Same-PreProc.  & 38.7 & 72.5 & 53.1 \\
Senet50 & Same-PreProc.  &  69.6 & 88.1 & 84.8 & Inception & Same-PreProc.  & 23.9 & 59.5 & 44.3  \\
VGG16  & Same-PreProc.  & 66.5 & 83.2 & 72.6  & Xception  & Same-PreProc.  & 25.5 & 64.6 & 47.7  \\
Facenet & Same-PreProc.  & 48.4 & 12.8 & 5.6 & ResNetV2 & Same-PreProc.  & 32.3 & 67.7 & 49.4 \\
OpenFace  & Same-PreProc.  & 33.5 & 35.1 & 34.9 & DenseNet  & Same-PreProc.  & 24.8 & 67.8 & 46.2 \\ \midrule

Resnet50 & Diff-PreProc. & 15.1 & 25.3 & 17.0 & ResNet & Diff-PreProc. &   3.51 &  25.4 &  21.3 \\
Senet50 & Diff-PreProc. & 32.9 & 41.2 & 23.3 & Inception & Diff-PreProc. & 2.39 & 14.8 &  14.7 \\
VGG16  & Diff-PreProc. & 28.8 & 39.3 & 31.9 & Xception  & Diff-PreProc. &  1.18 &  21.9 &  21.2  \\
Facenet & Diff-PreProc. & 21.1 & 9.9 & 5.3 & ResNetV2 & Diff-PreProc. & 2.33 & 23.5 &  23.1  \\
OpenFace  & Diff-PreProc. & 13.3 & 32.3 & 31.1  & DenseNet  & Diff-PreProc. & 2.92 & 16.7 & 15.2  \\ 
\bottomrule
\end{tabular}
}
\label{tab:impdata}
\end{table*}

\subsection{Adversarial Impersonation}
\label{sub:adv-imper}
In this section, we assess if the biometric reconstructions can effectively gain authentication on the target biometric system.
Specifically, we apply the reconstructions to the two biometric tasks described in Section~\ref{sec:setup}: \textit{verification} and \textit{identification}. For the verification task, we evaluate the attack success rate in two cases: \textit{Same-Image}, comparing the reconstruction against the original image; \textit{Same-Subject}, comparing the reconstruction with a different image from the same subject.
In practice, Same-Image is the case of an attacker who stole the \textit{stored} template used for verification, while Same-Subject reflects the case of an attacker intercepting the \textit{new} feature vector. The results, reported in Table~\ref{tab:impdata}, demonstrate that our biometric reconstructions are effective for impersonation. In particular, up to 92\% of face (SeNet50) and 73\% of fingerprint (ResNet) reconstructions pass \textit{verification}, while up to 73\% (SeNet50) and 42\% (ResNet) of crafted face and fingerprint samples, respectively, receive correct identification. In Appendix~\ref{app:mul-emb}, we also discuss the case in which the attacker has multiple embeddings from the same subject to cater for biometric systems that use more than one embedding per user.

\descr{Impact of attacker's data.} In Table~\ref{tab:impdata}, we also observe that more restrictive conditions on the attacker's data affects adversarial impersonation. For example, face-identification accuracy ranges in [27\%, 70\%] under Same-PreProcessing conditions, but degrades to [13\%, 33\%] in the Diff-PreProcessing case. However, even reconstructions crafted under the most restrictive conditions pose considerable threat to the biometric systems. In \textit{Diff-PreProcessing}, face and fingerprint reconstructions still achieve 22\% and 19\% correct identifications, respectively, which is a gain of 73x and 60x compared to random/chance (0.3\%).

\begin{figure}[t]
 \centering
 \includegraphics[width=1\columnwidth]{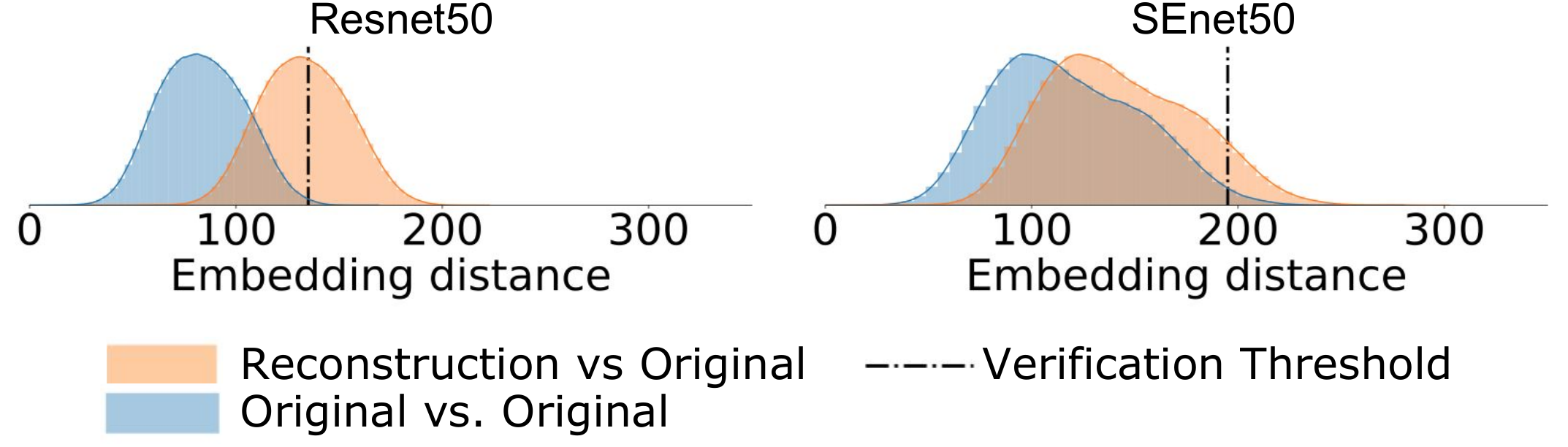}
\caption{Distance between original and reconstruction embeddings in ResNet50 and SeNet50 target models. ResNet50 is more resilient against impersonation with face reconstructions, as shown by the larger distances between the embeddings.}
 \label{fig:senet}
\end{figure}

\descr{Impact of the DNN.} The attacker's success rates are also strongly dependent on the target DNN model, especially for face reconstructions. In particular, the case of Facenet is interesting again, which yields the smallest reconstruction errors (based on Table~\ref{tab:recmodels}), but also the worst success rates on impersonation.
Perhaps more importantly, we observe that ResNet50 and SeNet50 (or ResNet and Inception) provide very different success rates in impersonation, despite being similarly trained (Table~\ref{tab:targets}) and yielding similar reconstruction errors (based on Table~\ref{tab:recmodels}). For instance, in the case of \textit{Same-Identity} attacker's data, the TAR for reconstructions is 92\% for SeNet50 and ``only'' 45\% for ResNet50. Compared to SeNet50, ResNet50 detects adversarial reconstructions better, as shown in Figure~\ref{fig:senet} by the different distances between original and reconstruction embeddings.

\descr{Impact of adaptations on the feature extractor.} 
Table~\ref{tab:impft} reports the impersonation results in the case of Fine-Tuning (FT) of the target-model under same pre-processing. As expected, progressive fine-tuning leads to reconstructions that are less effective for impersonation. This reflects the increasing reconstruction quality degradation observed in Figure~\ref{fig:ftex}. However, even in the FT-5 case, where the target model is additionally trained on a large dataset of $\ge$ 50K images, the impersonation success rates are non negligible: based on TAR values, 3 to 5 \% reconstructions still get accepted. Interestingly, for more \textit{moderate} setups FT-1 and FT-2 (fine tuning on 10K-20K new samples), the reconstructions maintain at least 50\% of their effectiveness compared to No-Adapt case, both in verification and identification. Table~\ref{tab:impft} also shows the results under Diff. PreProc. conditions on the attacker's data. This is the most challenging case for the attacker, but success rates are still significantly higher than random-chance. 

\descr{Impact of the number of embeddings.} The attack success rates strongly increase if the attacker has access to multiple biometric embeddings for the target identity\footnote{For example, maliciously obtained from an authentication system that stores multiple templates of a subject, like TouchID~\cite{touch-id}}. As shown in Table~\ref{tab:mul-emb}, with 5 embeddings, the attack performance for the \textit{Diff. PreProcrocessing} case increases from 15\% to 22.8\% in Identification (+54\% identification accuracy), and from 17\% to 38\% in Authentication (+121\% TAR).

\begin{table}[t]
\caption{Adversarial impersonation results when applying fine-tuning (FT) on the target model (ResNet50 for faces, ResNet for fingerprints). Random-chance identification accuracy is 0.3\% for both biometric modalities.}
\centering
\begin{tabular}{llll}
\toprule
\textbf{Target model's} & \textbf{Identification} &  \multicolumn{2}{c}{\textbf{TAR @ 1\% FAR}}  \\ 
\textbf{adaptation} & \textbf{accuracy [\%]} & \textbf{Same-Image} & \textbf{Same-Subject} \\ \midrule
\multicolumn{2}{l}{\textit{\textbf{Same PreProcessing:}}} \\
& Face / Fing. & Face / Fing. & Face / Fing. \\
~~No-Adapt & 29.2 / 38.5 & 38.1 / 71.8 & 33.3 / 52.9\\
~~Adapt:FT-1 & 19.4 / 36.7 & 36.5 / 65.7 & 29.9 / 45.2 \\
~~Adapt:FT-2 & 18.2 / 35.5 & 35.1 / 54.8 & 30.1 / 34.3\\
~~Adapt:FT-3 & 11.7 / 22.7 & 15.5 / 46.0 & 14.6 / 33.6\\
~~Adapt:FT-4 & 8.83 / 15.3 & 12.8 / 34.7 & 13.9 / 31.4\\
~~Adapt:FT-5 & 3.42 / 4.15 & 7.13 / 10.1 & 5.60 / 8.30\\
\midrule
\multicolumn{2}{l}{\textit{\textbf{Diff. PreProcessing:}}} \\
& Face / Fing. & Face / Fing. & Face / Fing. \\
~~No-Adapt & 15.1 / 3.51 & 25.4 / 25.4 & 20.2 / 21.3\\
~~Adapt:FT-1 & 12.3 / 3.04 & 21.3 / 21.6 & 19.8 / 20.1\\
~~Adapt:FT-2 & 10.1 / 2.83 & 18.2 / 20.8 & 14.2 / 18.9\\
~~Adapt:FT-3 & 6.3 / 2.45 & 10.5 / 9.3 & 9.70 / 8.5\\
~~Adapt:FT-4 & 3.4 / 1.73 & 7.7 / 5.02 & 11.1 / 4.32\\
~~Adapt:FT-5 & 0.9 / 1.51 & 2.5 / 4.44 & 1.6 / 3.9 \\
\midrule
\end{tabular}
\label{tab:impft}
\end{table}

\subsection{Comparison with Non-Adversarial Embedding Inversion}

We have compared our results with the ones obtained using the non-adversarial embedding inversion approach by Dosovitskiy et al. in \cite{dosovitskiy2015inverting}. With access to the original data and to all layers of the original model (both missing in the adversarial settings), the approach in \cite{dosovitskiy2015inverting} trains an inversion network minimizing the image-space L2 loss. In \cite{dosovitskiy2015inverting}, the lowest reconstruction errors are obtained when the inversion network is trained together with the original model in the auto-encoder setup. We adopt this case as our baseline, which provides an upper-bound to the reconstruction quality. To enable the comparison with our approach, we implemented the baseline on ResNet50 face recognition model. 

\begin{figure}[t]
  \includegraphics[width=1\columnwidth]{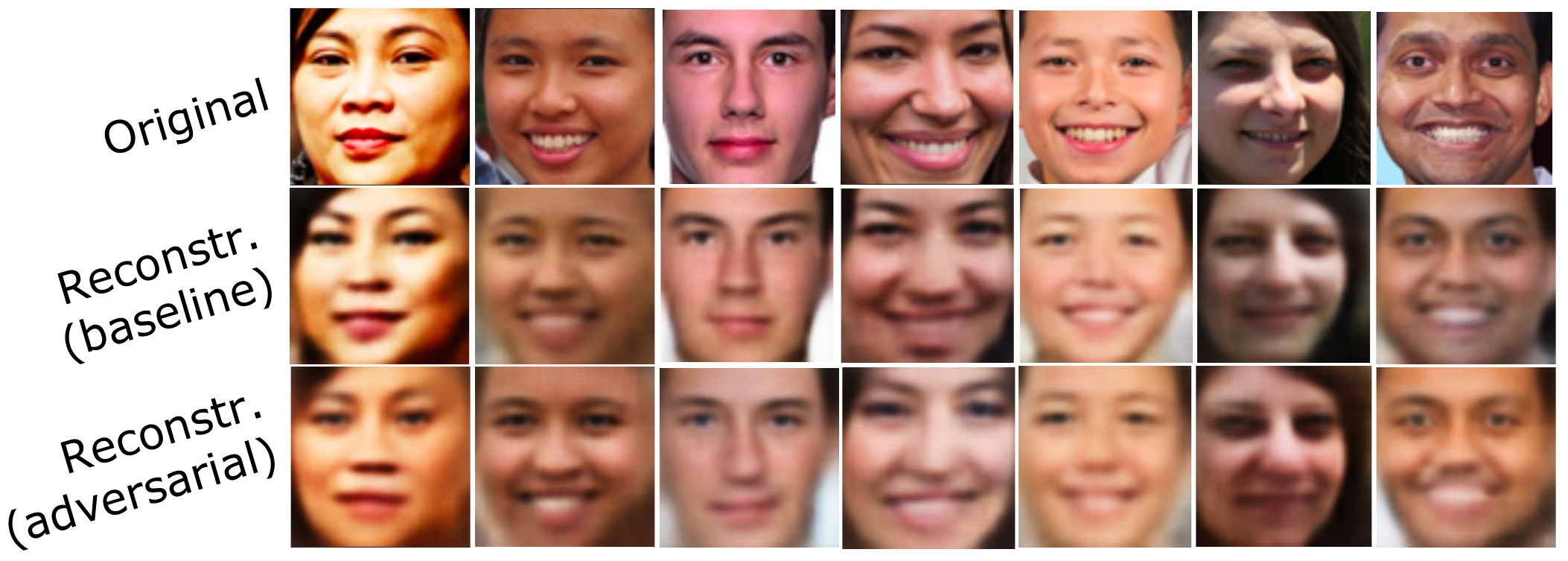}
\caption{Comparison between our face reconstructions (adversarial) and the ones obtained with Dosovitskiy et al. non-adversarial approach~\cite{dosovitskiy2015inverting} (baseline). In both cases, the target model is ResNet50.}
\label{fig:baseline}
\end{figure}

As shown in Figure~\ref{fig:baseline}, despite the adversarial conditions, our attack can craft reconstructions that are close to the baseline ones. In addition, Table~\ref{tab:compimp} reports the adversarial impersonation results obtained with the baseline non-adversarial inversion. We observe that the setup in \cite{dosovitskiy2015inverting} yields higher success rates: 52.2\% in identification, 69.3\% in Same-Image verification, and 48.0\% in Same-Subject verification. However, with respect to the non-adversarial results, our reconstructions still exhibit up to 56\% of the effectiveness in identification and up to 69\% of the effectiveness in Same-Subject verification.

\begin{table}[t]
\caption{Adversarial impersonation results: comparison between our face reconstructions (Adversarial) and the ones obtained with Dosovitskiy et al. non-adversarial approach~\cite{dosovitskiy2015inverting} (Baseline). In both cases, the target model is ResNet50.}
\centering
\begin{tabular}{llll}
\toprule
\textbf{Target model's} & \textbf{Identification} &  \multicolumn{2}{c}{\textbf{TAR @ 1\% FAR}}  \\ 
\textbf{adaptation} & \textbf{accuracy [\%]} & \textbf{Same-Image} & \textbf{Same-Subject} \\ \midrule
Baseline~\cite{dosovitskiy2015inverting} & 52.2 & 69.3 & 48.0 \\ [1.5mm]
Adversarial & 29.2 & 38.1 & 33.3 \\
(No-Adapt) \\ [1.5mm]
Adversarial & 18.2 & 35.1 & 30.1 \\
(FT-2) \\
\midrule
\end{tabular}
\label{tab:compimp}
\end{table}

\begin{table}[h]
\caption{Adversarial impersonation results when the attacker has access to multiple (N) face embeddings for the target identity. For Identification, the identity prediction is obtained by \textit{ensemble}: out of the N reconstructions, the attacker keeps the identity predicted on the highest number of reconstructions. For Verification TAR, we measure the chance of the attacker gaining authentication from at least one of the reconstructions. The target model is ResNet50.}
\label{tab:mul-emb}
\centering
\begin{tabular}{llll}
\toprule
$N$ & Attacker's & Identification &  TAR @ 1\% FAR \\ 
 & data & accuracy [\%] & (Same-Subject) \\ \midrule

1 & Same-Ident. & 28.1 & 41.7  \\
5 & Same-Ident. & 45.2 & 71.2   \\
10 & Same-Ident. & 60.4 &  74.5 \\
\midrule

1 & Same-PreProc. & 27.2  & 29.8 \\
5 & Same-PreProc. & 31.9  & 44.4 \\
10 & Same-PreProc.  &  39.6  & 46.7  \\
\midrule

1 & Diff-PreProc. & 15.1 & 17.0 \\
5 & Diff-PreProc. & 22.8 & 37.7 \\
10 & Diff-PreProc. & 35.3 & 43.9 \\
\bottomrule
\end{tabular}
\end{table}

\section{Inversion When Target Pre-trained DNN is Inaccessible}
\label{sec:lim}

In Section~\ref{sec::advscenario}, we have assumed that the biometric recognition system builds on a pre-trained model from a prominent architecture, available in open model repositories (e.g., \cite{tfhub}~\cite{modelzoo}~\cite{onnx}). After creating an extensive collection of available pre-trained models, the attacker finds which one has probably generated the observed feature vector. In the case of target model adaptations (fine-tuning), the attacker can still partially recover the target model {\em as long as} the original DNN falls within the attacker's collection. However, there are cases where the initial DNN is inaccessible to the attacker, if the original pre-trained model is acquired (e.g., purchased) from a private source, or if the target model is built and trained from scratch using a large-scale dataset. Building a new model ``from scratch'' is not an option for several biometric modalities (iris~\cite{iris}, palmprint~\cite{palmprint}, ear~\cite{ear}, fingerprint~\cite{fingernet}~\cite{michelsanti2017fast}), but it is reasonable for face recognition where massive open datasets like VGGFace2~\cite{vggface2} exist. 

We now relax the assumption on the target model provenance and we investigate the reconstructions obtained when the target model (or any initial version of it) is not in the attacker's pool. To experiment with this condition, we fix the model returned by Model Inference (we set $\hat{\Phi}$ equal to ResNet50 from Table~\ref{tab:targets}) and we vary the real target model used by the biometric recognition system. Specifically, we choose 4 pre-trained models from different DNN architectures but all providing 2048-long feature vectors: ResNet50V2~\cite{kerasapp}, ResNeXt~\cite{resnext}, DenseNet~\cite{kerasapp}, SeNet50~\cite{kerasvggface}. We additionally train these models on a face dataset (Facescrub). Upon obtaining a feature vector from one of these models and before giving it to the reconstructor, we min-max normalize each of the 2048 embedding dimensions according to the numerical ranges of the embeddings from $\hat{\Phi}$. 

\begin{figure}[t]
  \includegraphics[width=1\columnwidth]{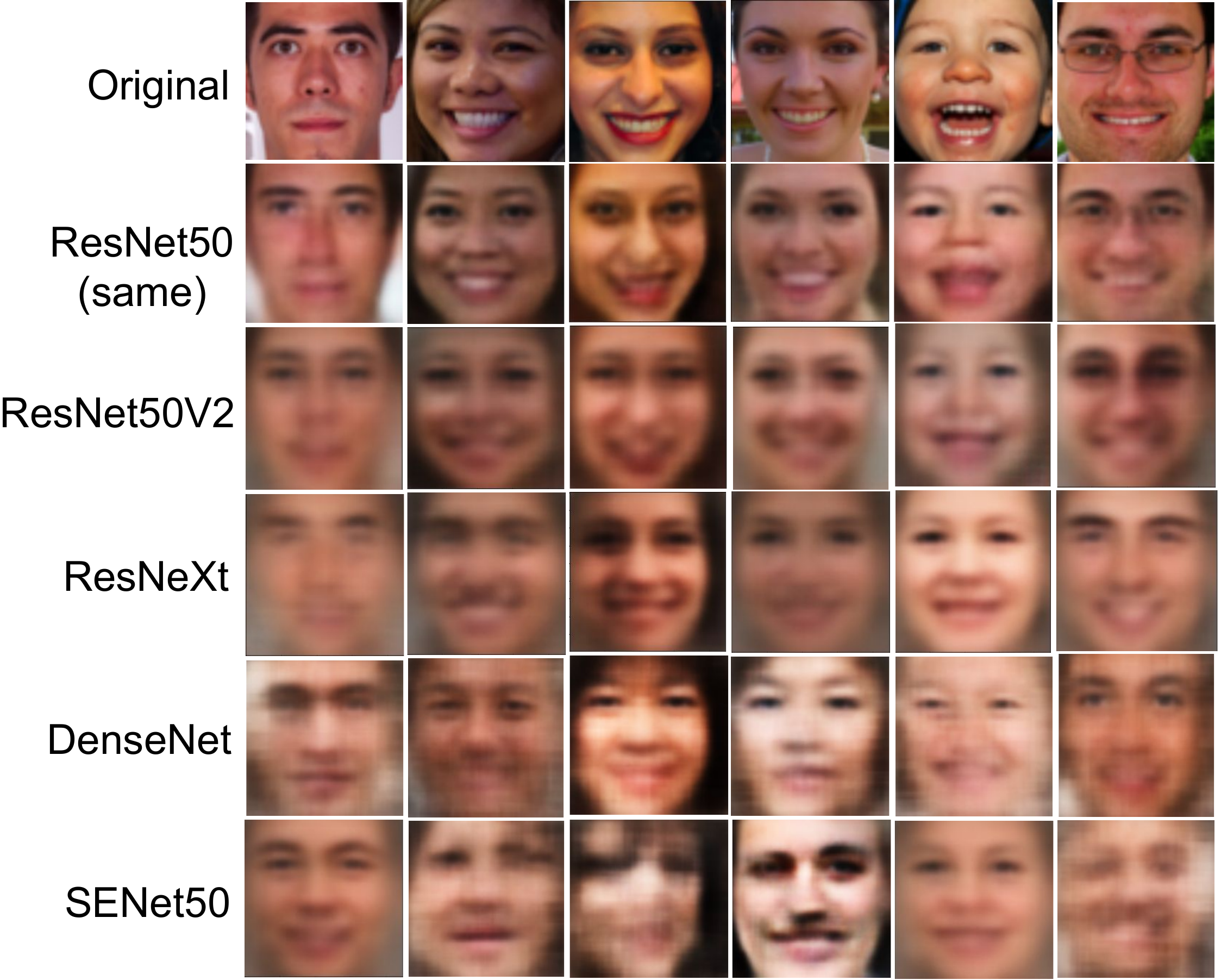}
\caption{Face reconstructions obtained when the target model (or any initial version of it) is not in the attacker's collection. We have fixed the model used to train the face reconstructor (= ResNet50), while varying the real target model.}
\label{fig:crossmodel}
\end{figure}

Figure~\ref{fig:crossmodel} shows that the difference between the attacker's inferred model $\hat{\Phi}$ and the real target model can significantly deteriorates the reconstruction quality. The misalignment between the latent spaces of the two DNNs results in loss of sharpness, obfuscations and, in some cases (DenseNet and SeNet50 target models), even identity and gender mismatch. This observation is confirmed by the adversarial impersonation results in Table~\ref{tab:limits}. For example, for SeNet50 target model and under the Same-PreProc. condition on the attacker's data, ``only'' 4.12\% of reconstructions are correctly classified. This is still not negligible (a 14x gain compared to random-choice), but much lower compared to what obtained if the target model is in the attacker's pool (27.2\%). 

At the same time, we observe that in the case of ResNet50V2 target model (which is not in the attacker's pool), the reconstructions maintain core features of shape, size and orientation, as well as partial face semantics. These reconstructions also maintain up to 65\% of the effectiveness in adversarial impersonation compared to the more favorable case of ResNet50 (which instead is in the attacker's pool). This is surprising, and can be explained by the stronger similarity between ResNet50 and ResNet50V2 in terms of their architectures\footnote{The two DNNs have similar architectures. ResNet50V2 differs from the original ResNet50 by including \textit{pre-activations}~\cite{he2016identity} in the residual units.}, with respect to the other experimented models. Overall, these results show that our attack can at least partially recover the original biometric trait even when the real target model is not in the attacker's pool, as long as the attacker has access to a ``similar'' DNN, and with minimal assumptions on the original biometric data.  

\begin{table}[ht]
\caption{Adversarial impersonation results obtained when the target model (or any initial version of it) is not in the attacker's collection. We have fixed the model used to train the face reconstructor (= ResNet50), while varying the real target model.  Random-chance identification accuracy is 0.3\%.}
\centering
\begin{tabular}{lllll}
\toprule

\textbf{Target} & \textbf{Attacker's} & \textbf{Identif.} &  \multicolumn{2}{c}{\textbf{TAR @ 1\% FAR}} \\ 
\textbf{model} & \textbf{data} & \textbf{acc.[\%]} & \textbf{Same-Img.} & \textbf{Same-Subj.} \\ \midrule

{\color{gray} ResNet50} & {\color{gray} Same-Ident.} & {\color{gray}28.1} & {\color{gray}45.4} & {\color{gray}41.7}\\
ResNet50V2  & Same-Ident. & 18.20 & 25.07 & 16.34  \\
ResNeXt & Same-Ident. & 13.43 & 18.01 & 12.20  \\
DenseNet & Same-Ident. & 7.86 & 15.76 & 10.09  \\
SENet50  & Same-Ident. & 4.94 & 9.01 & 5.52 \\ \midrule

{\color{gray}ResNet50} & {\color{gray}Same-PreProc.} & {\color{gray}27.2} & {\color{gray}32.0} & {\color{gray}29.8} \\
ResNet50V2  & Same-PreProc. & 12.95 & 15.53 & 11.05  \\
ResNeXt & Same-PreProc. & 9.35 & 13.54 & 9.89  \\
DenseNet & Same-PreProc. & 6.05 & 11.98 & 9.16  \\
SENet50  & Same-PreProc. & 4.12 & 6.88 & 3.81 \\ \midrule

{\color{gray}ResNet50} & {\color{gray}Diff-PreProc.} & {\color{gray}15.1} & {\color{gray}25.3} & {\color{gray}17.0} \\
ResNeXt & Diff-PreProc. & 8.75 & 12.13 & 9.46  \\
ResNet50V2  & Diff-PreProc. & 4.94 & 8.83 & 6.16  \\
DenseNet & Diff-PreProc. & 2.2 & 6.96 & 4.65  \\
SENet50  & Diff-PreProc. & 1.01 & 2.95 & 2.57 \\ 


\bottomrule
\end{tabular}
\label{tab:limits}
\end{table}

\vspace{-0.5cm}

\section{Conclusion}
We have investigated the inversion of deep biometric embeddings under {\em adversarial} settings, where an attacker has only access to a feature-space representation.
We proposed an embedding inversion attack which builds on the capability to (partially) infer the original DNN model from its \textit{footprint} on feature vectors. 

Extensive experiments with two prominent biometric modalities (\textit{face} and \textit{fingerprint}) show that the attack provides accurate predictions about the target model, and it yields effective reconstructions even when (i) the attacker has no access to similarly pre-processed biometric data; (ii) the attacker can only approximate the target DNN as this was re-purposed through additional training. 
We observed that reconstructions can significantly lose effectiveness if the target model is additionally trained on large-scale data, e.g., $\ge 50\text{K}$. 
However, it is worth noting that such conditions are unlikely, as in practice the pre-trained DNNs are generally re-purposed on \textit{moderate} datasets~\cite{survey2019}, e.g., limited samples in ear recognition~\cite{ear}, limited samples per identity in fingerprint recognition~\cite{fingernet}. 

 Our tests on adversarial impersonation also showed that different DNN models may exhibit different levels of resilience against reconstructions. This may happen, as shown by the case of ResNet50 and SENet50, even when the DNN models are trained with the same datasets, have similar recognition performance, generate same-length embedding, and are tested with reconstructions of same quality. This shows that there is also an ``architectural" factor that makes DNN models more or less vulnerable to adversarial impersonation. We plan to further explore such relations in future work, in addition to extending our experimentation to other biometric modalities. 
 

\bibliographystyle{plain}
\bibliography{ccs-sample}


\appendix
\section{Attacker's Data Preprocessing}
\label{appendix:alignments}
\textit{Face:} ~To set the \textit{Diff-PreProcessing} condition on face images, we have tested the attack on face images that are differently {\em aligned} compared to the attacker's ones. Alignment refers to a combination of several face image attributes: tightness of crop around the face, relative position of eyes, nose, mouth, ears in the image (face keypoints). We have experimented with two different alignment techniques, namely Dlib~\cite{dlib} and MTCNN~\cite{mtcnn}.

\begin{itemize}
    \item Dlib: The pre-trained Dlib face detector~\cite{dlib} is used to detect the position of face keypoints in the image. A set of landmark points are pre-defined for a particular image size, which correspond to the expected location of each facial keypoint in the image. An affine transformation is then used to normalize the image such that the facial keypoints are present at the expected locations in the image. Based on the landmark positions, a bounding box is obtained which is used to obtain a tight crop on the face.
    \item MTCNN: The Multi-Task Cascaded Convolutional Network (MTCNN)~\cite{mtcnn} implements a three stage proposal, refinement and regression methodology to obtain the bounding box and face keypoints. This information is used to obtain a tight crop on the face. MTCNN is the primary technique our attacker uses to align the face images prior to reconstructor training and embedding extraction.
\end{itemize}

\begin{figure}[h]
\centering
  \includegraphics[scale=0.60]{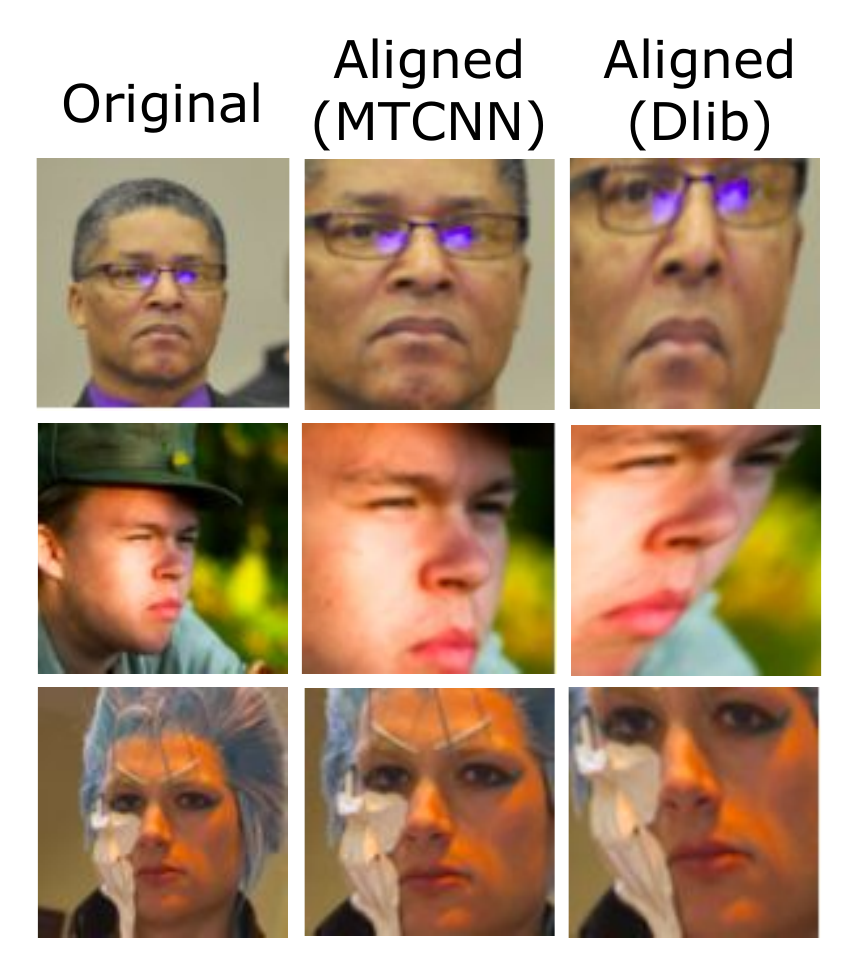}
\caption{Original images and cropped-aligned images for the case of DLib and MTCNN alignment techniques.}
\label{fig:alignments}
\end{figure}

\noindent The primary difference between the two techniques is that DLib alignment enforces the same position for facial keypoints using an affine transformation on the original image, prior to applying the bounding box crop. In MTCNN, the bounding box crop is applied directly based on the positions of facial keypoints in the original image. We provide an example of the two face alignment techniques in Figure~\ref{fig:alignments}.  

\begin{figure}[h]
\centering
 \includegraphics[scale=0.65]{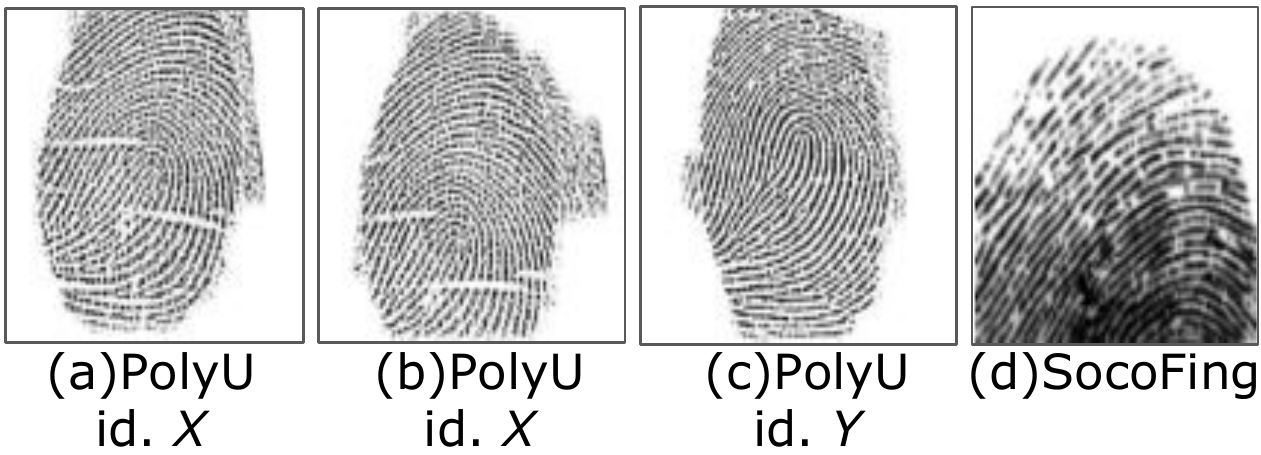}
\caption{Fingerprint images (a) and (b) are different images belonging to the same identity in  PolyU dataset. Fingerprint image (c) belongs to a different identity from PolyU dataset. Fingerprint image (d) is a fingerprint sample from SocoFing dataset.} 
\label{fig:fingdatasets}
\end{figure}

\textit{Fingerprint:} ~To set the \textit{Diff-PreProcessing} condition for fingerprints, we tested the attack on fingerprints from a different dataset. Figure~\ref{fig:fingdatasets} shows through examples the difference between the datasets, which is likely the result of different sensor type, and specifications of the acquisition device.

\section{Embedding Extraction}
\label{appendix:mi}
In Table~\ref{tab:embed-descript}, we detail the different embedding-extraction configurations  used in Model Inference experiments (Section~\ref{sec:miresults}) and to evaluate the biometric reconstruction quality for different embedding lengths (Figure~\ref{fig:emblen}).

\begin{table}[h]
\caption{Embedding extraction configurations.}
\centering
\begin{tabular}{ lll }
\toprule
\multicolumn{3}{c}{\textbf{Fingerprint embedding}} \\ \midrule
Model id & Extraction Layer & Embedding Size\\ \midrule
ResNet & conv5\_block3\_avg\_pool  & 2048\\
ResNetV2 & conv5\_block3\_avg\_pool  & 2048\\
Inception &  global\_average\_pooling2d\_7 & 2048\\
Xception  & global\_average\_pooling2d\_6  & 2048\\ \midrule
DenseNet  & final\_avg\_pool & 1024\\
ResNet  & conv5\_block1 & 1024\\
Xception  & block13\_pool & 1024 \\
ResNetV2 & conv5\_block3\_2\_relu  & 1024\\ \midrule
Inception  & global\_average\_pooling2d\_6  & 512\\
Xception   &global\_average\_pooling2d\_5 & 512\\
ResNet  & conv5\_block3\_2\_relu & 512\\ 
ResNetV2  & conv5\_block3\_2\_relu & 512\\ \midrule
\multicolumn{3}{c}{\textbf{Face embedding}} \\ \midrule
Model id & Extraction Layer & Embedding Size\\ \midrule
ResNet50 & activation\_49 & 2048 \\
SENet50 & activation\_130 & 2048 \\
Facenet & Block\_35\_1\_activation & 2048 \\
OpenFace & Inception\_5b\_1\_act & 2048 \\ \midrule
VGG16 & pool5 & 512 \\
Facenet &  Bottleneck\_BatchNorm & 512\\
ResNet50   & conv3\_1\_1x1\_reduce & 512 \\
SENet50  &  conv5\_1\_1x1\_reduce & 512\\ \midrule
FaceNet & final\_norm\_layer & 128\\
OpenFace &  final\_norm\_layer & 128\\
ResNet50 & conv3\_4\_1x1\_reduce & 128\\
SENet50 & conv5\_1\_1x1\_down & 128\\ \bottomrule
\end{tabular}
\label{tab:embed-descript}
\end{table}

\newpage
\section{Reconstruction Models}
\label{app:rec}
In Table~\ref{tab:reconstructors}, we detail the architecture of the reconstruction models for different input (\textit{i.e.}, embedding) size.  We implemented all reconstruction models using Keras, and trained them for 200 epochs with batch size 16 and Adagrad optimizer. This took for each model roughly 10 to 12 hours on one NVIDIA Tesla P100 GPU card. 

\begin{table}[h]
\caption{Architectures of our reconstruction models for different input embedding lengths. Batch Normalization is used after every Convolutional and Transpose Convolutional Layer before applying activation. In the table, K=Kernel, S=Stride, D=Depth, U=Units.}
\centering
\begin{tabular}{ lllll }
\toprule
\multicolumn{5}{c}{\textbf{Reconstructor: input embedding of length 2048}} \\ \midrule
Layer Type & Parameters & In.Shape & Out.Shape & Activation\\ \midrule
Dense & U:16384 & 2048 & 16384 & Leaky ReLU\\
Reshape & - & 16384 & (4,4,1024) & -\\
Conv2D\_Transpose & K:5,S:2,D:512 & (4,4,1024) & (8,8,512) & Leaky ReLU\\
Conv2D\_Transpose & K:5,S:2,D:256 & (8,8,512) & (16,16,256) & Leaky ReLU\\
Conv2D\_Transpose & K:5,S:2,D:128 & (16,16,256) & (32,32,128) & Leaky ReLU\\
Conv2D\_Transpose & K:5,S:2,D:3 & (32,32,128) & (64,64,3) & -\\
\midrule
\multicolumn{5}{c}{\textbf{Reconstructor: input embedding of length 1024}} \\ \midrule
Layer Type & Parameters & In.Shape & Out.Shape & Activation\\ \midrule
Dense & U:16384 & 1024 & 16384 & Leaky ReLU\\
Reshape & - & 16384 & (4,4,1024) & -\\
Conv2D\_Transpose & K:5,S:2,D:512 & (4,4,1024) & (8,8,512) & Leaky ReLU\\
Conv2D\_Transpose & K:5,S:2,D:256 & (8,8,512) & (16,16,256) & Leaky ReLU\\
Conv2D\_Transpose & K:5,S:2,D:128 & (16,16,256) & (32,32,128) & Leaky ReLU\\
Conv2D\_Transpose & K:5,S:2,D:3 & (32,32,128) & (64,64,3) & -\\
\midrule
\multicolumn{5}{c}{\textbf{Reconstructor: input embedding of length 512}} \\ \midrule
Layer Type & Parameters & In.Shape & Out.Shape & Activation\\ \midrule
Dense & U:32768 & 512 & 32768 & Leaky ReLU\\
Reshape & - & 32768 & (8,8,512) & -\\
Conv2D\_Transpose & K:5,S:2,D:256 & (8,8,512) & (16,16,256) & Leaky ReLU\\
Conv2D\_Transpose & K:5,S:2,D:128 & (16,16,256) & (32,32,128) & Leaky ReLU\\
Conv2D\_Transpose & K:5,S:2,D:3 & (32,32,128) & (64,64,3) & -\\
\midrule
\multicolumn{5}{c}{\textbf{Reconstructor: input embedding of length 128}} \\ \midrule
Layer Type & Parameters & In.Shape & Out.Shape & Activation\\ \midrule
Dense & U:4096 & 128 & 4096 & Leaky ReLU\\
Reshape & - & 4096 & (4,4,256) & -\\
Conv2D\_Transpose & K:5,S:2,D:128 & (4,4,256) & (8,8,128) & Leaky ReLU\\
Conv2D\_Transpose & K:5,S:2,D:64 & (8,8,128) & (16,16,64) & Leaky ReLU\\
Conv2D\_Transpose & K:5,S:2,D:64 & (16,16,64) & (32,32,32) & Leaky ReLU\\
Conv2D\_Transpose & K:5,S:2,D:3 & (32,32,32) & (64,64,3) & -\\
\bottomrule
\end{tabular}
\label{tab:reconstructors}
\end{table}

\section{Fine-Tuning of Target Model}
\label{app:fttl}
Table~\ref{tab:face-deg-across} describes in detail the fine-tuning (FT) of \textit{ResNet50} face recognition model on \textit{Facescrub} dataset. To allow for fine-tuning with larger datasets, we follow standard tranfer-learning practices and ``open" a larger portion of the original DNN model for additional training. All FT configurations have validation accuracy $\ge95\%$.

\begin{table*}[!ht]
\caption{\small{FT configurations on ResNet50 face recognition model classification model.
All FT configurations have validation accuracy $\ge95\%$.} \label{tab:ftconfigs}}
\resizebox{\linewidth}{!}{%
\begin{tabular}{llllllllllll}
\toprule
\textbf{FT Configuration} & FT Dataset & \# new identities & \# new samples & \multicolumn{6}{c}{Additional training epochs}\\
 & & & &  Dense & Block\_5 & Block\_4 & Block\_3 & Block\_2 & Block\_1 \\
\midrule
No-Adapt  & - & - & - & 0 & 0 & 0 & 0 & 0 & 0\\
FT-1 & Facescrub & 64 & 8079 & 1 & 15 & 0 & 0 & 0 & 0\\
FT-2 & Facescrub & 179 & 15005 & 1 & 10 & 15 & 0 & 0 & 0\\
FT-3 & Facescrub & 223 & 23343 & 1 & 10 & 10 & 15 & 0 & 0\\
FT-4 & Facescrub & 288 & 34548 & 1 & 10 & 10 & 25 & 0 & 0\\
FT-5 & Facescrub & 377 & 58878 & 1 & 10 & 10 & 10 & 10 & 15\\
\bottomrule
\end{tabular}
}
\label{tab:face-deg-across}
\end{table*}


\section{Lipschitz Constant of Reconstruction Network}
\label{app:lip}
Inspired on the spectral analysis of neural network instability in \cite{szegedy2013intriguing}, we represent the reconstruction model (which is, in essence, a DNN) as a sequence of $K$ layers, each with Lipschitz constant $L_k$:
\begin{equation}
        \forall x,\delta : ||\phi_k(x, W_k) - \phi_k(x+\delta, W_k)|| \le L_k||\delta||
\end{equation}
where $W_k$ represent the parameters/weights of the k-th layer ($\phi_k$) of the reconstruction network. We can obtain the Lipschitz constant of the reconstructor by composing the model layers:
\begin{equation}
      L \le \prod_{k=1}^K{L_k}
\end{equation}
Since pooling and normalization layers are \textit{contractive}~\cite{szegedy2013intriguing}~\cite{gouk2018regularisation}, a conservative value for $L$ can be obtained by considering $L_k$ for activation, dense (fully-connected) layers, and transpose convolution layers:

\begin{itemize}
    \item \textit{Activation}: $L_k=1$ for many activations types, including ReLu and Leaky ReLu (the one used in our reconstructors)~\cite{miyato2018spectral};
    \item \textit{Dense}: For a dense layer, the upper-bound $L_k$ is the spectral norm of its weights $||W||$, equal to the largest singular value of the fully connected matrix~\cite{szegedy2013intriguing}~\cite{miyato2018spectral}. We compute it based on the implementation in \cite{lipfc}.
    \item \textit{Conv2D\_transpose}: For a transpose convolution layer, we build on the observation that Conv2D\_transpose can be expressed as a Conv2D operation performed using the same filter, but on 0-padded input. To compute $L_k$ for the Conv2D operation, which is the spectral norm of its weights, we rely on recent work on the recent analysis in \cite{sedghi2018singular} (and its implementation~\cite{convsv}).
\end{itemize}




\section{Access to Multiple Embeddings}
\label{app:mul-emb}
Some biometric authentication systems use multiple templates, where a sample is compared with multiple templates. For the attacker, having $N$ embeddings would make adversarial impersonation easier. To show this, we repeat the experiments from Section~\ref{sub:adv-imper} for face recognition, but this time varying the number of embeddings $N$ of the subject, where $N \in \{1, 5, 10\}$. Not surprisingly, as the number of embeddings increases, the identification accuracy and true acceptance rates (at 1\% FAR) increase, showing substantial improvement over the default case of $N = 1$ (see Table~\ref{tab:impdata}). These results are summarised in Table~\ref{tab:mul-emb}.

\end{document}